%% file: icml2026.tex
\documentclass{article}

\usepackage{microtype}
\usepackage{graphicx}
\usepackage{subcaption}
\usepackage{booktabs}
\usepackage{hyperref}

\usepackage[utf8]{inputenc}
\usepackage[T1]{fontenc}
\usepackage{url}
\usepackage{amsfonts}
\usepackage{nicefrac}
\usepackage{xcolor}

\usepackage{adjustbox}
\usepackage{array}
\usepackage{mathrsfs}
\usepackage{tabularx}
\usepackage{multirow}
\usepackage{rotating}
\usepackage{float}
\usepackage{newfloat}
\usepackage{listings}

\usepackage{algorithm}
\usepackage{algorithmic}

\usepackage[preprint]{icml2026}

\usepackage{amsmath}
\usepackage{amssymb}
\usepackage{mathtools}
\usepackage{amsthm}
\usepackage[capitalize,noabbrev]{cleveref}

\theoremstyle{plain}
\newtheorem{theorem}{Theorem}[section]

\newtheorem{lemma}[theorem]{Lemma}
\newtheorem{corollary}[theorem]{Corollary}
\theoremstyle{definition}
\newtheorem{definition}[theorem]{Definition}

\theoremstyle{remark}

\newcommand{\squishlisttwo}{
   \begin{list}{$\bullet$}
       { \setlength{\itemsep}{0pt}      \setlength{\parsep}{3pt}
       \setlength{\topsep}{3pt}       \setlength{\partopsep}{0pt}
      \setlength{\leftmargin}{1.5em} \setlength{\labelwidth}{1em}
       \setlength{\labelsep}{0.5em} } }

\DeclareMathOperator*{\E}{\mathbb{E}}

\begin{document}

\twocolumn[
  \icmltitle{Learning Over Dirty Data with Minimal Repairs}


  \begin{icmlauthorlist}
    \icmlauthor{Cheng Zhen}{inst1}
    \icmlauthor{Prayoga}{inst1}
    \icmlauthor{
Nischal Aryal}{inst1}
    \icmlauthor{
Arash Termehchy}{inst1}
    \icmlauthor{Garrett Biwer}{inst1}
    \icmlauthor{
Lubna Alzamil}{inst1}
  \end{icmlauthorlist}

  \icmlaffiliation{inst1}{Oregon State University}

  \icmlcorrespondingauthor{Cheng Zhen}{zhenc@oregonstate.edu}

  \icmlkeywords{Machine Learning, ICML, Missing Data}

  \vskip 0.3in
]

\printAffiliationsAndNotice{} 

\begin{abstract}
Missing data often exists in real-world datasets, requiring significant time and effort for data repair to learn accurate models. In this paper, we show that imputing all missing values is not always necessary to achieve an accurate ML model. We introduce concepts of minimal and almost minimal repair, which are subsets of missing data items in training data whose imputation delivers accurate and reasonably accurate models, respectively. Imputing these subsets can significantly reduce the time, computational resources, and manual effort required for learning. We show that finding these subsets is NP-hard for some popular models and propose efficient approximation algorithms for wide range of models. Our extensive experiments indicate that our proposed algorithms can substantially reduce the time and effort required to learn on incomplete datasets.
\end{abstract}

\input{introduction}
\input{background}
\input{minimalRepair}

\input{MinimalForACM}

\input{experimentalEvaluation}

\input{related}
\section*{Impact Statement}
This work aims to reduce the cost and effort of data preparation by identifying only essential missing values to repair, which can streamline machine learning pipelines and improve accessibility in resource-constrained settings. At the same time, in high-stakes applications such as healthcare or criminal justice, a lack of domain validation could lead to missed critical repairs and biased for unsafe models. These risks are not unique to our approach and can be mitigated through domain expertise and standard model validation practices prior to deployment.

\bibliography{ICML2026}
\bibliographystyle{icml2026}

\newpage
\appendix
\onecolumn
\input{supp_MR_for_linear_regression}
\input{supp_constrained_optimization_AMR}
\input{supp_experimental_setting}

\input{supp_experimental_result}

\input{supp_robustness}
\input{supp_additional_related_and_code}

\input{supp_proof}

\end{document}

%% file: introduction.tex
\section{Introduction}
\label{sec:introduction}
The performance of an ML model is highly dependent on the quality of its training data. 
In real-world data, a major data quality issue is missing or incomplete data \cite{Emmanuel21,10.1145/3035918.3054775,krishnan2016activeclean,10.1145/2882903.2912574,10.1145/3447548.3470817}. 
There are two common approaches to address missing values in training data. 
The first approach involves deleting samples with missing values. 
However, this method can lead to the loss of important information and introduce bias \cite{van2018flexible}. 
Another popular approach is data repair or imputation, in which end-users or ML practitioners impute missing values with the correct ones \cite{andridge2010review,farhangfar2008impact,NEURIPS2021_5fe8fdc7,little2002statistical,smieja2018processing,pelckmans2005handling,whang2023data,williams2005incomplete}. 
Accurate repair is often challenging and expensive as it usually requires extensive collaboration with expensive domain experts. 
It usually must be repeated whenever the dataset evolves.

To reduce the cost of imputation, significant effort has been made to train imputation models on the observed subset of the dataset that predict accurate values for missing data items \cite{10.18637/jss.v045.i03,10.5555/3540261.3542084,NEURIPS2021_5fe8fdc7,10.1093/bioinformatics/btr597,pmlr-v80-yoon18a,zheng2022diffusion}.
State-of-the-art models for data imputation may take a long time to process and predict values for missing data items, and those that use deep neural networks need costly computational resources \cite{perini2024database,zheng2022diffusion,pmlr-v80-yoon18a}.
As the dataset evolves, the user often has to repeat these steps.
Moreover, in domains where important decisions must be made, e.g., healthcare and criminal justice, humans may need to manually verify the predictions of the imputation models \cite{10.14778/1952376.1952378}. 
Some users also distrust black-box model-based imputation techniques in critical applications and prefer to reason about missing data themselves using observed features and domain knowledge \cite{ahmad2019challengeimputationexplainableartificial,stempfle2025handlingmissingvaluesclinical}. 
In addition, model-based imputation may perform poorly when the ratio of missing data to observed data is too large \cite{doi:10.1177/2192568218811922,2017/12/06,Junaid25}. 
In these settings, users may have to manually repair at least parts of the data.

To address these challenges, we introduce the concept of a {\bf minimal repair} for a training dataset with missing values. 
Generally speaking, this subset represents the smallest group of data items with missing values that, once repaired, yields the same model as that trained on a fully and accurately repaired dataset. 
By finding and imputing this set, users can significantly reduce the time and effort required to manually repair a dataset without sacrificing model accuracy.
It also reduces the time and computational resources needed to predict missing values using imputation models and the manual labor required to verify their imputations. 
Moreover, minimal repair of a dataset pinpoints the subset of the dataset whose uncertainty impacts the effectiveness of the model trained on the dataset.
Hence, it simplifies the inspection and debugging of model training, which is often labor intensive \cite{10.1145/3617338}. 
Because incomplete data sets are prevalent and often evolve, a small reduction in time, effort, and computational resources in the preparation of training datasets can save significant resources in the long run. Specifically, our contributions are as follows.

\begin{itemize}
\item We define minimal repair for learning support vector machines (SVM) (Section \ref{sec:minimalSVM}) and linear regression (Appendix \ref{sec:supp_MR_for_linear_regression}) over incomplete data. We prove that finding minimal repairs for SVM and linear regression is NP-hard and propose efficient algorithms with provable error bounds to approximate minimal repairs for them.

\item Minimal repairs may sometimes be too large or take too long to find. 
We propose the concept of \textbf{almost minimal repair}, which is the minimal subset of data items with missing values whose repair delivers a model with a loss within a given threshold from the model trained over the fully and accurately repaired dataset.
We prove that the problem of finding almost minimal repairs is NP-hard for SVM and linear regression. We propose algorithms with provable error bounds to approximate almost minimal repairs for models with convex loss. We also propose algorithms to approximate almost minimal repairs for neural networks (Section \ref{section:AMR}).  

\item We evaluated the scalability of our algorithms on multiple real-world datasets (Section \ref{sec:experimental-evaluation}).
Our empirical results indicate that our proposed algorithms efficiently approximate minimal and almost minimal repairs and deliver models with the same or almost the same accuracy as those trained over fully repaired datasets. 
Our results also indicate that using minimal and almost minimal repairs can reduce the time of the model-based imputation methods for large data often without losing accuracy in the downstream learning task.
\end{itemize}

%% file: background.tex
\section{Background}
\label{sec:Background}
We model the training data as a table where each row represents a training sample. 
One column in the table represents labels and others represent the features of the samples. 
Given that the training data has $d$ features, we denote its features as $[\mathbf{z}_1, \dots, \mathbf{z}_d]$.
The values of each feature belong to the {\it domain of the feature}, e.g., real numbers.
To simplify our analysis, we assume that all the features share the same domain. 
Our results extend to other settings.
A {\it training set} with $n$ samples is a pair of a feature matrix $\mathbf{X} =[\textbf{x}_{1}, ..., \textbf{x}_{n}]^{T}$ and a corresponding label vector $\textbf{y} = [y_{1}, ..., y_{n}]^{T}$. 
We denote each sample with $d$ features in $\textbf{X}$ as a vector $\textbf{x}_i = [x_{i1}, ..., x_{id}]$, where $x_{ij}$ represents the $j^{th}$ feature in the $i^{th}$ sample.
Given the training set $(\mathbf{X},\mathbf{y})$, the target function $f$, and the loss function $L$, the goal of training is to find an optimal model $\mathbf{w}^* = \arg \min\limits_{\mathbf{w}\in \mathcal{W}} L(f(\mathbf{X},\mathbf{w}),\mathbf{y})$. 

\paragraph{{\bf Missing values}} 
Any $x_{ij}$ is a missing value if it is unknown (marked by \textit{null}). 
An {\it incomplete sample} ({\it incomplete feature}) is a sample (feature) with at least one missing value.
We use {\it complete feature} and {\it complete sample} to refer to features and samples that are free of missing values. 
We denote the set of all missing values in a feature matrix $\mathbf{X}$ as $M(\mathbf{X})$, the set of incomplete samples as $MS(\mathbf{X})$, and the set of incomplete features as $MF(\mathbf{X})$.
In this paper, we focus on the case where all missing values are in the feature matrix and the label vector is complete.
 
\paragraph{{\bf Repair}}
A repair is a complete version of an incomplete feature matrix $\textbf{X}$ where all missing values in $\textbf{X}$ are replaced with values from their domains and the complete values of $\textbf{X}$ remain intact.  
Given the repair $\textbf{X}^r$ of the feature matrix $\textbf{X}$, we denote the repair, i.e. imputation, of the sample $\textbf{x}_i$ in $\textbf{X}$ by $\textbf{x}_i^r$.
Since the domains of features often contain numerous or infinite values, an incomplete feature matrix usually has many or infinitely many repairs. We denote this set of all repairs of $\textbf{X}$ by $\textbf{X}^{R}$.

%% file: minimalRepair.tex
\section{Minimal Repair (MR)}
\label{sec:minimalSVM}
We use the concept of \textit{certain model} \cite{zhen2024certain} to define minimal repair. 
A model $\textbf{w}^*$ is a certain model for the target function $f$ on the training set $(\mathbf{X}, \mathbf{y})$ if for every repair $\textbf{X}^{r} \in \textbf{X}^{R}$, we have $\textbf{w}^*$ $=$ $\mathop{\arg \min}\limits_{\textbf{w}\in \mathcal{W}} L(f(\textbf{X}^{r}, \textbf{w}), \textbf{y})$ where $L$ is the loss function.
Intuitively, a certain model minimizes training loss for all repairs of the incomplete feature matrix.
Thus, if a certain model exists, one can learn an accurate model over the training set without any repair to the training data, as training over any repair to the dataset, e.g., using randomly selected values, will deliver the same accurate model. This observation holds regardless of the missingness mechanism—Missing Completely at Random (MCAR), Missing at Random (MAR), or Missing Not at Random (MNAR).
Given the restrictive definition of certain models, they do not often exist \cite{zhen2024certain}.
Thus, we find the minimal repair of an incomplete training set such that the resulting training set has a certain model.

\begin{definition}\label{definition:minimalImputation} 
A set of incomplete samples $\mathbf{S}_{MR}$ in the training set $(\mathbf{X}, \mathbf{y})$ 
is a minimal repair for learning a target model if we have: 1) a certain model exists when imputing all missing values in $\mathbf{S}_{MR}$, and 2) there is no other set $\mathbf{S'}$ satisfying condition (1) such that $|\mathbf{S'}| < |\mathbf{S}_{MR}|$ where $|\mathbf{S}|$ denotes the cardinality of the set $|\mathbf{S}|$.
\end{definition}

According to Definition~\ref{definition:minimalImputation}, the certain model exists regardless of the imputed values in $\mathbf{S}_{MR}$. 
Obviously, if these values are accurate, e.g., by using experts or advanced imputation models, or inaccurate, the certain model may be accurate or inaccurate, respectively.
Our aim is \textit{not} to improve the accuracy of imputations directly, but to reduce resources used for such imputations using minimal repairs. We also discuss the robustness of models from minimal repairs in Appendix \ref{sec:supp_robustness}.

Since the concept of a certain model relies on exact optimality, finding the minimal repair generally requires analytical solutions that are model-specific. In this paper, we explore minimal repair for two ML models: SVM and linear regression. We focus on SVM in this section and present the approaches and experimental results for linear regression in the Appendix \ref{sec:supp_MR_for_linear_regression} and \ref{sec:supp_experimental_result}, respectively.

\subsection{Finding Minimal Repair for SVM}

We denote the minimal repair for SVM with the regularization parameter $C$ on the training set $(\mathbf{X}, \mathbf{y})$ as $\mathbf{S}_{MR}(\mathbf{X}, \mathbf{y},C)$.
We have the following property regarding its uniqueness.

\begin{theorem}\label{theorem:SVMminImputeUnique} 
Given training set $(\mathbf{X}, \mathbf{y})$ and regularization parameter $C$, $\mathbf{S}_{MR}(\mathbf{X}, \mathbf{y},C)$ is unique.
\end{theorem}

Let $SV(\mathbf{X^r}, \mathbf{y}, C)$ be the set of support vectors for the optimal SVM model with regularization parameter $C$ on a repair $\mathbf{X^r}$ of the training set ($\mathbf{X}, \mathbf{y}$). 

\begin{lemma}\label{theorem:SVMminImputeEquivalence} 
 Given the training set $(\mathbf{X}, \mathbf{y})$ and the regularization parameter $C$, at least one repair $\mathbf{x}^r_i$ of every sample $\mathbf{x}_i \in$ $S_{MR}(\mathbf{X}, \mathbf{y}, C)$ is a support vector in a repair $\mathbf{X^r}$ of $\mathbf{X}$, i.e., $\mathbf{x}^r_i \in SV(\mathbf{X^r}, \mathbf{y}, C)$.
\end{lemma}

Hence, to determine if an incomplete sample belongs to the minimal repair, one could materialize every repair of the feature matrix and check if the incomplete sample is a support vector for any of them. 
However, this process can be extremely inefficient due to the often large number of repairs. 
Assume that each missing value $x_{ij}$ is bounded by an interval $[x_{ij}^{min}, x_{ij}^{max}]$ based on its domain.
$\mathbf{X}^e$ is an {\it edge repair} to $\mathbf{X}$ if for every missing value $x_{ij}$, $x_{ij}^e = x_{ij}^{min}$ or $x_{ij}^{max}$. 
$\mathbf{X}^E$ denotes the set of all edge repairs for $\mathbf{X}$. 
Theorem \ref{theorem:SVMonlycheckedgerepairs} shows that we can use only the edge repair instead of all repairs to check if an incomplete sample belongs to the minimal repair.

\begin{theorem}\label{theorem:SVMonlycheckedgerepairs}
Given the training set $(\mathbf{X}, \mathbf{y})$ and the regularization parameter $C$, an incomplete sample $\mathbf{x}_i$ belongs to minimal repair $S_{MR}(\mathbf{X}, \mathbf{y}, C)$ if and only if there is at least one edge repair $\mathbf{X}^e$ of $\mathbf{X}$ such that $\mathbf{x}_i^e \in SV(\mathbf{X}^e, \mathbf{y}, C)$ where $\mathbf{x}_i^e$ is the repair of $\mathbf{x}_i$.
\end{theorem}
Based on Theorem \ref{theorem:SVMonlycheckedgerepairs}, we can find the minimal repair following these steps: 1) Initialize an empty minimal repair set, $S_{MR}$. 2) Iterate over each incomplete sample $\mathbf{x}_i$. At each iteration, materialize all edge repairs $\mathbf{X}^e \in \mathbf{X}^E$, and check if $\mathbf{x}_i$ is a support vector for any of the edge repairs. 
If it is, add $\mathbf{x}_i$ to $S_{MR}$, and 3) Finally, return the minimal repair $S_{MR}$.
Despite this optimization, finding the minimal repair remains computationally intractable.

\begin{theorem}\label{theorem:SVMNPHard}
Given a training set $(\mathbf{X}, \mathbf{y})$ with missing values, deciding whether an incomplete sample 
belongs to the minimal repair for SVM on $(\mathbf{X}, \mathbf{y})$ is NP-hard. Consequently, finding the minimal repair for SVM is NP-hard.
\end{theorem}
 
\subsection{Approximating Minimal Repair}
\label{sec:approx-minimal-repair}

We propose an efficient approximation algorithm (Algorithm~\ref{alg:SVMapproximate}) to find minimal repair for SVM. 
Its key idea is to test whether each incomplete sample $\mathbf{x}_i$ belongs to minimal repair by constructing an edge repair $\mathbf{X}^e$ that maximizes the likelihood of $\mathbf{x}_i$ becoming a support vector.
This construction begins with a random edge repair and iteratively updates each missing value in the dataset to its minimum or maximum bound. 
At each step, this choice minimizes $y_i \mathbf{w}^\top \mathbf{x}_i$, encouraging $\mathbf{x}_i$ to satisfy the support vector condition $y_i \mathbf{w}^\top \mathbf{x}_i \le 1$. 
If this condition holds after the full pass of the data, $\mathbf{x}_i$ is selected for repair.

Crucially, this algorithm \textbf{does not return any false positive}. Since the algorithm initializes with a randomly selected edge repair, it does not introduce bias towards any specific imputation in learning models.

\begin{theorem}\label{theorem:SVMapproximatingneveroverimpute}
Every sample returned by Algorithm~\ref{alg:SVMapproximate} belongs to $S_{MR}(\mathbf{X}, \mathbf{y}, C)$.
\end{theorem}

\begin{algorithm}[!htbp]
\footnotesize
\caption{Approximating minimal repair for SVM on training set $(\mathbf{X}, \mathbf{y})$}\label{alg:SVMapproximate}
\begin{algorithmic}
\STATE $S_{MR} \gets \emptyset$

\STATE $\mathbf{X}^e \gets$ a random edge repair to the feature matrix $\mathbf{X}$
\FOR{$\mathbf{x}_i \in MS(\mathbf{X})$}
    \FOR {$x_{pq} \in M(\mathbf{X})$}
        \STATE $\mathbf{X}^{e_{min}}, \mathbf{X}^{e_{max}} \gets$ two edge repairs by only replacing $x_{pq}$ in $\mathbf{X}^e$ with its min or max value
        \STATE $\mathbf{w}_1, \mathbf{w}_2 \gets SVM(\mathbf{X}^{e_{min}}, \mathbf{y}), SVM(\mathbf{X}^{e_{max}}, \mathbf{y})$         \COMMENT{learning SVM models with edge repairs}
       \STATE $\mathbf{X}^e \gets
\begin{cases}
\mathbf{X}^{e_{\min}}, &
y_i \mathbf{w}_1^\top \mathbf{x}_i^{e_{\min}}
\le
y_i \mathbf{w}_2^\top \mathbf{x}_i^{e_{\max}} \\
\mathbf{X}^{e_{\max}}, & \text{otherwise}
\end{cases}$

    \ENDFOR
    \STATE $\mathbf{w} \gets SVM(\mathbf{X}^e, \mathbf{y})$
    \STATE \textbf{if } $y_i \mathbf{w}^\top \mathbf{x}_i \leq 1$ \textbf{ then } $S_{MR} \gets S_{MR}.add(\mathbf{x}_i)$
\ENDFOR
\STATE return $S_{MR}$
\end{algorithmic}
\end{algorithm}

Since each iteration modifies only one missing value, adjacent models $\mathbf{w}_1$ and $\mathbf{w}_2$ differ by only a single feature entry. This allows us to avoid retraining from scratch by applying incremental or decremental SVM updates~\cite{cauwenberghs2000incremental,laskov2006incremental}. These techniques update the model efficiently---typically an order of magnitude faster---by reusing computations from the previous solution.

Algorithm~\ref{alg:SVMapproximate} may miss some samples of minimal repair. 
Thus, we iteratively apply Algorithm~\ref{alg:SVMapproximate} to the remaining incomplete samples in the training set to find more samples in the minimal repair of the training set. The process ends when no new samples are selected for repair.
The following theorem shows that the probability of not finding samples of minimal repair decreases using this approach.

\begin{theorem}\label{theorem:SVMapproximatingrate2}
Given the training set $(\mathbf{X}, \mathbf{y})$, let $p_k(\mathbf{x})$ be the probability that an incomplete sample $\mathbf{x}$ in minimal repair of $(\mathbf{X}, \mathbf{y})$ not returned in iteration of $k>0$ in iterative application of Algorithm~\ref{alg:SVMapproximate}, $p_k(\mathbf{x})$ $>$ $p_{k+1}(\mathbf{x})$.
\end{theorem}

\begin{corollary}
If the probability distribution of each missing value is known, and we let $g(x_{ij})$ denote the probability density function of the ground truth value for the missing value $x_{ij}$ in the incomplete training set $(\mathbf{X},\mathbf{y})$. 
If missing values in $\mathbf{X}$ are independent, the probability that an incomplete sample $\mathbf{x}_i$ in minimal repair  not returned by Algorithm 1 in the main content is:
\begin{equation}
p(\mathbf{x}_i) = 1- \frac{\idotsint_{\min( x_{ij}^{\text{visited}})}^{\max(x_{ij}^{\text{visited}})}
\prod_{x_{ij} \in M(\mathbf{X})} g(x_{ij})\,dx_{ij}}
{\idotsint_{x_{ij} \in M(\mathbf{X})} \prod_{x_{ij} \in M(\mathbf{X})} g(x_{ij}) \,dx_{ij}}
\end{equation}

$x_{ij}^{\text{visited}} \in$ $\{x_{ij}^{min}, x_{ij}^{max}\}$ shows the values used for $x_{ij}$ in Algorithm \ref{alg:SVMapproximate}.
\end{corollary}

%% file: MinimalForACM.tex
\section{Almost Minimal Repair}
\label{section:AMR}

Minimal repair guarantees that the learned model exactly achieves the optimal loss that would be obtained from the fully repaired dataset. However, enforcing exact optimality often requires model-specific analytical solutions (as discussed in Section \ref{sec:minimalSVM}) and may yield repair sets that are too large or computationally expensive to compute. 
In practice, particularly with optimization methods like Stochastic Gradient Descent (SGD), finding the exact global optimum is often not necessary; instead, people typically aim for a model whose loss is sufficiently close to the optimal loss.

To align with this practical reality and reduce repair costs, we relax the definition of minimal repair. We aim for a set of incomplete samples whose imputation allows learning a model that is \textit{near-optimal}—within a specific loss threshold—for all possible repairs. This relaxation mirrors the flexibility of numerical optimization: just as SGD converges to a neighborhood of the optimum, AMR seeks a repair that guarantees the model lands within a neighborhood of a certain model. We use the concept of approximately certain model (ACM) \cite{zhen2024certain} to formalize this notion. 
For a user-defined threshold $e \geq 0$, $\mathbf{w}^{\approx}$ is an ACM for target function $f$ on training set $(\mathbf{X}, \mathbf{y})$ if for every repair $\mathbf{X}^r$,  
$L(\mathbf{w}^{\approx}, \mathbf{X}^r, \mathbf{y})$ $-$ $\min_{\mathbf{w} \in \mathcal{W}} L(\mathbf{w}, \mathbf{X}^r, \mathbf{y})$ $\le e$.

\begin{definition}
Given a threshold $e \ge 0$, a set $S_{\text{AMR}}$ of incomplete samples in the training set $(\mathbf{X}, \mathbf{y})$ is an almost minimal repair (AMR) for the target function $f$ with loss $L$ if: (1) repairing $S_{\text{AMR}}$ yields an ACM for $f$ in $(\mathbf{X}, \mathbf{y})$, and (2) no other set $S'$ satisfies (1) with $|S'| < |S_{\text{AMR}}|$.
\end{definition}

If $e = 0$, ACM reduces to a certain model. Hence, finding the exact AMR is at least as hard as finding the exact MR. As established previously, computing MR is NP-hard for SVM and linear regression; consequently, finding the exact AMR is also NP-hard (details in the Appendix).

In this section, we focus on developing algorithms for approximating AMR. We first show the approach for \textbf{models with convex loss functions}. Under convexity, finding a solution sufficiently close to the global optimum is generally tractable. This allows us to bound the optimality gap and provide theoretical guarantees on the repair set. Then, we show the challenge of finding AMR for \textbf{neural networks with non-convex loss functions}, and the approach to approximate AMR with relaxation of the AMR definition.

\subsection{Approximating AMR for models with convex losses}
\label{sec:amr_algorithm}

We formulate the AMR problem as a constrained optimization task. Let $\mathbf{s} \in [0, 1]^m$ be the vector of continuous repair scores for $m$ incomplete samples. We define the robust suboptimality gap $g(\mathbf{w}, \mathbf{s})$ as the maximum possible difference between the model's loss and the optimal loss under any valid repair consistent with $\mathbf{s}$:
\begin{equation}
    g(\mathbf{w}, \mathbf{s}) = \max_{\mathbf{X}^r} \left( L(\mathbf{w}, \mathbf{X}^r(\mathbf{s}), \mathbf{y}) - \min_{\mathbf{v} \in \mathcal{W}} L(\mathbf{v}, \mathbf{X}^r(\mathbf{s}), \mathbf{y}) \right)
\end{equation}
where $\mathbf{X}^r(\mathbf{s})$ denotes a repair where samples with $s_i=1$ are fixed to their ground truth (simulated or oracle) and samples with $s_i < 1$ are chosen adversarially. Consistent with Theorem \ref{theorem:SVMonlycheckedgerepairs}, for convex loss functions, the worst-case repair $\mathbf{X}^r$ lies among the \textit{edge repairs}. Thus, the adversary searches the space of edge repairs to maximize this gap.

Our objective is to minimize the repair mass subject to the ACM constraint:
\begin{equation}\label{eq:AMR_primal_dual}
    \min_{\mathbf{s} \in [0,1]^m, \mathbf{w} \in \mathcal{W}} \sum_{i} s_i \quad \text{subject to} \quad g(\mathbf{w}, \mathbf{s}) \le e
\end{equation}

We solve this using a Primal-Dual approach on the Lagrangian $\mathcal{L}(\mathbf{s}, \mathbf{w}, \lambda) = \sum_{i} s_i + \lambda (g(\mathbf{w}, \mathbf{s}) - e)$. Algorithm \ref{alg:AMR_PrimalDual} details the procedure. In each iteration, we perform an \textbf{Adversarial Step} to find the current worst-case edge repair $\mathbf{X}_{adv}$. We then update the model $\mathbf{w}$ (Primal Step) to minimize the loss on this repair. Simultaneously, we perform a \textbf{Repair Update} on $\mathbf{s}$ using gradient descent on the Lagrangian; if a sample contributes to the violation of the threshold $e$, the term $\lambda \nabla_s g$ forces $s_i$ to increase. Finally, the \textbf{Dual Update} adjusts $\lambda$ based on the constraint violation.

\begin{algorithm}[!htbp]
\footnotesize
\caption{Primal-Dual Approximation for AMR}\label{alg:AMR_PrimalDual}
\begin{algorithmic}
\STATE \textbf{Input:} Training set $(\mathbf{X}, \mathbf{y})$, threshold $e$, learning rates $\eta_w, \eta_s, \eta_\lambda$
\STATE \textbf{Initialize:} Repair scores $\mathbf{s} \gets \mathbf{0}$, Model $\mathbf{w}$ randomly, Dual $\lambda \gets 1$
\STATE $\mathbf{X}_{adv} \gets$ Random edge repair
\WHILE{not converged}
    \STATE \COMMENT{\textbf{Step 1: Adversarial Search (Edge Repair)}}
    \STATE Find edge repair $\mathbf{X}_{adv}$ that maximizes the gap $g(\mathbf{w}, \mathbf{s})$
    \STATE \COMMENT{\textbf{Step 2: Primal Model Update}}
    \STATE $\mathbf{w} \gets \mathbf{w} - \eta_w \nabla_{\mathbf{w}} \mathcal{L}(\mathbf{s}, \mathbf{w}, \lambda; \mathbf{X}_{adv})$
    \STATE \COMMENT{\textbf{Step 3: Repair Score Update}}
    \STATE $\mathbf{s} \gets \text{Proj}_{[0,1]} \left( \mathbf{s} - \eta_s \nabla_{\mathbf{s}} \mathcal{L}(\mathbf{s}, \mathbf{w}, \lambda; \mathbf{X}_{adv}) \right)$
    \STATE \COMMENT{\textbf{Step 4: Dual Variable Update}}
    \STATE $Gap_{curr} \gets L(\mathbf{w}, \mathbf{X}_{adv}, \mathbf{y}) - \min_{\mathbf{v}} L(\mathbf{v}, \mathbf{X}_{adv}, \mathbf{y})$
    \STATE $\lambda \gets \max(0, \lambda + \eta_\lambda (Gap_{curr} - e))$
\ENDWHILE
\STATE \textbf{Rounding:} Rank incomplete samples by scores $s_i$ and select the smallest set $S_{AMR}$ such that the ACM condition holds.
\STATE \textbf{Return:} $S_{AMR}$
\end{algorithmic}
\end{algorithm}

We now analyze the theoretical properties of Algorithm \ref{alg:AMR_PrimalDual}. Although the optimization problem in Equation \ref{eq:AMR_primal_dual} involves a non-convex constraint due to the difference-of-convex nature of the robust gap, we leverage the geometric properties of the loss function and the submodular structure of the repair problem to establish robust guarantees.

First, we address whether the algorithm can effectively find a valid repair and deliver ACM. The core requirement is that our repaired model must satisfy the ACM conditions, meaning the gap between our model's loss and the optimal loss must be small (within threshold $e$).

\begin{theorem}\label{theorem:ACM_converge}
Let $(\mathbf{w}_t, \mathbf{s}_t, \lambda_t)$ be the sequence generated by Algorithm \ref{alg:AMR_PrimalDual} at each iteration $t$. Assuming learning rates for $\mathbf{w}, \mathbf{s}$ and $\lambda$ that decrease over time to dampen stochastic noise (Robbins-Monro conditions: $\sum \eta_t = \infty, \sum \eta_t^2 < \infty$) \cite{robbins1951stochastic}, the solution converges to a feasible set that satisfies the ACM constraints in Equation \ref{eq:AMR_primal_dual}: $\limsup_{t \to \infty} [g(\bar{\mathbf{w}}_t, \bar{\mathbf{s}}_t) - e]^+ = 0$.
\end{theorem}

Intuitively, this result guarantees that the algorithm will eventually reach an ACM that satisfies the threshold $e$. The mechanism is mostly driven by the dual variable $\lambda$. If the robust gap is too large ($> e$), $\lambda$ grows, effectively increasing the penalty in the objective function. This forces the repair scores $\mathbf{s}$ to rise until the gap is reduced. Theorem \ref{theorem:ACM_converge} confirms that over enough iterations of running Algorithm \ref{alg:AMR_PrimalDual}, we achieve an ACM.

We also consider the efficiency of the algorithm. 

\begin{theorem}\label{theorem:ACM_convergence_rate}
The inner adversarial search of the worst-case repair $\mathbf{X}_{adv}$ (Step 1 of Algorithm \ref{alg:AMR_PrimalDual}) converges linearly in the case of piecewise linear-quadratic losses (e.g., Hinge loss), while the outer repair scores $\mathbf{s}$ (Step 3 of the algorithm) converge to a stationary point at a sublinear rate $O(1/\sqrt{T})$, where $T$ denotes the total number of iterations.
\end{theorem}

This theorem establishes that finding the worst-case repair in the algorithm is computationally efficient for models with piecewise linear-quadratic losses (e.g., linear SVM). The property of metric subregularity ensures that the inner maximization problem converges exponentially fast \cite{drusvyatskiy2018error}. While the outer optimization loop proceeds at a standard sublinear rate, the rapid convergence of the inner adversarial step ensures that the gradients driving the repair scores are computed efficiently, making the overall method scalable to large datasets.

Finally, since finding the exact AMR is NP-hard, our algorithm effectively performs a continuous version of a greedy search to approximate AMR.

\begin{theorem}\label{theorem:ACM_approx_bound}
The size of the repair set $S_{algo}$ output from Algorithm \ref{alg:AMR_PrimalDual} approximates the true minimal repair set $S_{opt}$ with a ratio bounded by a logarithmic factor: $|S_{algo}| \le O(\ln n) \cdot |S_{AMR}|$.
\end{theorem}

This theorem establishes a connection between our gradient-based update for $\mathbf{s}$ and a continuous relaxation of the greedy algorithm for the Set Cover problem \cite{chvatal1979greedy}. By iteratively increasing the score of the sample with the largest constraint violation—analogous to selecting the item with maximum marginal gain—our method achieves a logarithmic approximation ratio. This ensures that the AMR approximated by Algorithm \ref{alg:AMR_PrimalDual} has an upper bound on size that grows slowly with the number of samples in the dataset, $n$.

\subsection{Approximating AMR for Neural Networks}

The definition of ACM relies on the \textit{robust suboptimality gap}—the difference between the learned model's loss and the global optimal loss. Extending this definition to neural networks presents a fundamental challenge: the loss landscape of neural networks is non-convex, and the global optimal loss $\min_{\mathbf{w} \in \mathcal{W}} L(\mathbf{w}, \mathbf{X}^r, \mathbf{y})$ is generally unknown. Without access to the optimal loss, we cannot directly define ACM through the suboptimality gap or find AMR to achieve ACM. Furthermore, finding the worst-case repair in a non-convex inner maximization problem is itself challenging.

To address these challenges, we propose a relaxation of the AMR definition suitable for non-convex settings. Instead of requiring proximity to a global optimum, we seek a repair strategy that ensures the model converges to a \textit{Robust Stationary Point}. Specifically, we introduce the concept of \textbf{Robust Gradient Stationarity}.

In non-convex optimization, a standard criterion for convergence is reaching a stationary point where the gradient of the loss is sufficiently small. In the context of incomplete data, we want the model to be a stationary point for all repairs.

Let $\mathbf{s} \in [0, 1]^n$ be the repair selection vector as defined previously. We define the \textit{Robust Gradient Norm} $G(\mathbf{w}, \mathbf{s})$ as the maximum norm of the gradient over all valid repairs consistent with $\mathbf{s}$:

\begin{equation}
    G(\mathbf{w}, \mathbf{s}) = \max_{\mathbf{X}^r} \| \nabla_{\mathbf{w}} L(\mathbf{w}, \mathbf{X}^r(\mathbf{s}), \mathbf{y}) \|_2
\end{equation}

where $\mathbf{X}^r(\mathbf{s})$ represents the set of repairs where samples with $s_i=1$ are imputed, and samples with $s_i=0$ are not.

\begin{definition}
A model $\mathbf{w}$ is an $e$-Robust Stationary Point with respect to the repair selection $\mathbf{s}$ if the worst-case gradient magnitude is bounded across all possible imputations:
$G(\mathbf{w}, \mathbf{s}) \le e$.
\end{definition}

This definition implies that for the chosen repair set, regardless of the imputation results, the model is sufficiently close to convergence (i.e., having a small gradient). This serves as a proxy for the ACM condition in non-convex landscapes.

Since we cannot directly minimize the suboptimality gap, we adapt our optimization objective to minimize the repair cost subject to the robust stationarity constraint:
\begin{equation}
    \min_{\mathbf{s}, \mathbf{w}} \|\mathbf{s}\|_1 \quad \text{subject to} \quad G(\mathbf{w}, \mathbf{s}) \le e
\end{equation}

Solving this exactly is challenging due to the maximization of the gradient norm. We propose an iterative approximation algorithm that alternates between training the network, finding adversarial repair that maximizes the gradient norm, and selecting a subset of incomplete samples to repair and reduce this adversarial gradient.

\textbf{Adversarial Gradient Ascent:} Unlike the convex case where we searched for edge repairs to maximize loss, here we search for repairs that maximize the \textit{gradient norm}. This identifies the incomplete samples that prevent the optimizer from converging.

\textbf{Sensitivity Analysis:} To select repairs, we compute the sensitivity of the Robust Gradient Norm with respect to the uncertainty of each sample. Samples with high sensitivity—meaning their missing values allow for large fluctuations in the model's gradient—are prioritized for repair.

\begin{algorithm}[!htbp]
\footnotesize
\caption{Approximating AMR for Neural Networks}\label{alg:AMR_DNN}
\begin{algorithmic}
\STATE \textbf{Input:} Training set $(\mathbf{X}, \mathbf{y})$, threshold $\epsilon$, learning rates $\alpha, \beta$
\STATE \textbf{Initialize:} $\mathbf{w}$ randomly, Repair set $S_{AMR} \gets \emptyset$, $\mathbf{X}^{curr}$ imputed with mean
\WHILE{$G(\mathbf{w}, S_{AMR}) > e$}
    \STATE \COMMENT{\textbf{Step 1: Train Model on Current View}}
    \FOR{$k=1$ to $K$}
        \STATE $\mathbf{w} \gets \mathbf{w} - \alpha \nabla_{\mathbf{w}} L(\mathbf{w}, \mathbf{X}^{curr}, \mathbf{y})$
    \ENDFOR
    
    \STATE \COMMENT{\textbf{Step 2: Adversarial Search (Gradient Maximization)}}
    \STATE Find $\mathbf{X}_{adv}$ consistent with $S_{AMR}$ that maximizes $\|\nabla_{\mathbf{w}} L(\mathbf{w}, \mathbf{X}, \mathbf{y})\|$ using projected gradient ascent.
    
    \STATE \COMMENT{\textbf{Step 3: Active Constraint Identification}}
    \IF{$\|\nabla_{\mathbf{w}} L(\mathbf{w}, \mathbf{X}_{adv}, \mathbf{y})\| > \epsilon$}
        \STATE Compute sample-wise attribution of the gradient norm.
        \STATE Identify sample $i^*$ contributing most to the gradient variance.
        \STATE $S_{AMR} \gets S_{AMR} \cup \{i^*\}$
        \STATE \textbf{Repair:} Update $\mathbf{X}^{curr}_{i^*}$ with ground truth from oracle.
    \ELSE
        \STATE \textbf{Break} (Robust Stationarity Achieved)
    \ENDIF
\ENDWHILE
\STATE \textbf{Return:} $S_{AMR}$
\end{algorithmic}
\end{algorithm}

%% file: experimentalEvaluation.tex
\begin{table*}[!t]
\centering
\caption{Details of datasets with injected missing data}
\resizebox{\textwidth}{!}{%
\label{tab:synthetic-missing-charateristics}
\begin{tabular}{|c|c|c|c|c|c|}
\hline
Data Set & Task & Features & Training samples & Missing Factor\% & Missingness Type\\
\hline
{Malware} & Classification & 6823 & 1596 & 20-40-60 & MCAR, MAR, MNAR\\
\hline
{Tuadromd} & Classification & 242 & 3571 & 20-40-60 & MCAR, MAR, MNAR\\
\hline
{Credit Default} & Classification & 23 & 30000 & 20-40-60 &  MCAR, MAR, MNAR\\
\hline
{Gas} & Regression & 129 & 2566 & 20-40-60 & MCAR\\
\hline
{Superconductivity} & Regression & 82& 21262 & 20-40-60 &  MCAR\\
\hline
{Concrete} & Regression & 8 & 1030 & 20-40-60 & MCAR\\
\hline
\end{tabular}
}
\end{table*}

\section{Experimental Evaluation}\label{sec:experimental-evaluation}
\begin{table*}[!t]
\centering
\caption{ Details of datasets with original missing data}
\resizebox{\textwidth}{!}{%
\label{tab:real-world-missing-charateristics}
\begin{tabular}{|c|c|c|c|c|c|}
\hline
Data Set & Task & Features & Training samples & Missing Factor & Missingness Type\\
\hline
{Breast Cancer} & Classification & 10 & 559 & 1.97\% & MCAR\\
\hline
{Water-Potability} & Classification & 9 & 2620 & 39.00\% & MCAR\\
\hline
{Online-Ed} & Classification & 36 & 7026 & 35.48\% & MNAR, MCAR\\
\hline
{Bankruptcy} & Classification & 64 & 8402 & 54.00\% & MNAR \\
\hline
{Air Quality} & Regression & 12 & 7344 & 90.80\% & MNAR\\
\hline
{Cancer Rate} & Regression & 32 & 3048 & 81.00\% & MCAR \\
\hline
\end{tabular}
}
\end{table*}

\begin{table*}[!t]
\centering
\caption{Accuracy/time for SVM on data with injected MCAR}
\resizebox{\textwidth}{!}{%
\label{tab:injectedMVSVM}

\begin{tabular}{|c|c|c|ccc|ccc|ccc|}
\hline
\multirow{2}{*}{Data Set} & \multirow{2}{*}{\% Missing} & Ground Truth & \multicolumn{3}{c|}{Time(s)} & \multicolumn{3}{c|}{Accuracy(\%)} & \multicolumn{3}{c|}{Impute \% of Samples} \\
\cline{4-12} & & Accuracy(\%) &
AC & MR & AMR & AC & MR & AMR & AC & MR & AMR \\
\hline

\multirow{3}{*}{Malware} & 20 & 95.61 &
$\mathbf{1.36} \pm 0.15$ & $6.15 \pm 0.71$ & $2.2 \pm 0.2$ &
$93.13 \pm 0.76$ & $\mathbf{96.7} \pm 0.5$ & $95.6 \pm 0.3$ &
$6.39 \pm 0.54$ & $18.68 \pm 0.5$ & $\mathbf{1.9 \pm 0.8}$ \\
\cline{2-12}
& 40 & 95.03 &
$\mathbf{0.56} \pm 0.06$ & $9.8 \pm 0.92$ & $3 \pm 0.1$ &
$92.2 \pm 0.7^{*}$ & $92.42 \pm 3.5^{*}$ & $\mathbf{92.6 \pm 0.1}^{*}$ &
$3.35 \pm 0.46$ & $21.1 \pm 1.3$ & $\mathbf{2.1 \pm 0.4}$ \\
\cline{2-12}
& 60 & 95.91 &
$\mathbf{0.17} \pm 0.05$ & $12.81 \pm 1.05$ & $3.3 \pm 0.4$ &
$88.67 \pm 1.02$ & $\mathbf{96.37} \pm 0.7$ & $95.2 \pm 0.4$ &
$3.28 \pm 0.43^{*}$ & $16.65 \pm 0.7$ & $\mathbf{3 \pm 0.2}^{*}$ \\
\hline

\multirow{3}{*}{Tuadromd} & 20 & 98.67 &
${0.68 \pm 0.01}^{*}$ & $1.14 \pm 0.13$ & $\mathbf{0.6 \pm 0.2}^{*}$ &
$97.53 \pm 0.17$ & $\mathbf{98.73 \pm 0.1}$ & $97.8 \pm 0.1$ &
$3.78 \pm 0.48$ & $11.9 \pm 0.8$ & $\mathbf{1.9 \pm 0.1}$ \\
\cline{2-12}
& 40 & 98.77 &
$0.54 \pm 0.007$ & $2.19 \pm 0.34$ & $\mathbf{0.4 \pm 0.1}$ &
$97.42 \pm 0.13$ & $\mathbf{98.81 \pm 0.14}$ & $97.7 \pm 0.1$ &
$3.53 \pm 0.37$ & $11.1 \pm 1.26$ & $\mathbf{0.9 \pm 0.2}$ \\
\cline{2-12}
& 60 & 98.77 &
$\mathbf{0.34 \pm 0.004}^{*}$ & $3.29 \pm 0.16$ & ${0.4 \pm 0.1}^{*}$ &
$97.5 \pm 0.11$ & $\mathbf{98.77 \pm 0.1}$ & $97.9 \pm 0.2$ &
$2.48 \pm 0.24$ & $11.8 \pm 1.4$ & $\mathbf{1.7 \pm 0.5}$ \\
\hline

\multirow{3}{*}{Credit Default} & 20 & 81.03 &
$11.86 \pm 0.86$ & $\mathbf{1.39 \pm 0.08}$ & $5.6 \pm 2.3$ &
${81.02 \pm 0}^{*}$ & ${81.02 \pm 0.01}^{*}$ & $\mathbf{81.2 \pm 0.2}^{*}$ &
$0.19 \pm 0.03$ & $30 \pm 0$ & $\mathbf{0.1 \pm 0}$ \\
\cline{2-12}
& 40 & 81.03 &
$14.19 \pm 0.62$ & ${3.93 \pm 0.1}^{*}$ & $\mathbf{3.7 \pm 0.7}^{*}$ &
$81.02 \pm 0$ & $81 \pm 0.02$ & $\mathbf{81.3 \pm 0.2}$ &
$0.23 \pm 0.05$ & $30 \pm 0$ & $\mathbf{0.1 \pm 0}$ \\
\cline{2-12}
& 60 & 81.02 &
$14.2 \pm 0.77$ & $\mathbf{0.57 \pm 0.06}$ & $4.2 \pm 1.1$ &
$\mathbf{81.02 \pm 0}^{*}$ & $\mathbf{81.02 \pm 0.02}^{*}$ & $80.8 \pm 0.1$ &
$0.19 \pm 0.04$ & $30 \pm 0$ & $\mathbf{0.1 \pm 0}$ \\
\hline

\end{tabular}%
}
\end{table*}

\begin{table*}[!t]
\centering
\caption{Accuracy/time for SVM on data with original missingness using model-based imputation}
\resizebox{\textwidth}{!}{%
\label{tab:inherentMVSVM}

\begin{tabular}{|c|c|cccc|cccc|c|}
\hline
\multirow{2}{*}{Data Set} & \multirow{2}{*}{Method} & \multicolumn{4}{c|}{Time(s)} & \multicolumn{4}{c|}{Accuracy(\%)} & \multirow{2}{*}{\% Samples Imputed} \\
\cline{3-10}
 & & KNN & MICE & TCSDI & MF & KNN & MICE & TCSDI & MF &  \\
\hline
\multirow{4}{*}{Breast Cancer}
& MR & $0.066 \pm 0.003$ & $0.064 \pm 0.001$ & $51 \pm 3.4$ & $0.124 \pm 0.01$
& ${96 \pm 0.3}^{*}$ & ${96.1 \pm 0.7}^{*}$ & $97 \pm 0.5$ & $96.58 \pm 0.4$
& $18.18 \pm 1$ \\ \cline{2-11}

& AMR
& $0.006 \pm 0.001$ & $\mathbf{0.019 \pm 0.003}$ & $\mathbf{36 \pm 1}$ & $\mathbf{0.04 \pm 0}$
& ${\mathbf{96.1 \pm 0}}^{*}$ & ${96 \pm 0.1}^{*}$ & $97.1 \pm 0.1$ & $96.9 \pm 0$
& $\mathbf{6.7 \pm 0}$ \\ \cline{2-11}

& AC 
& $0.065 \pm 0.003$ & $0.065 \pm 0.002$ & $84 \pm 5$ & $0.12 \pm 0.01$
& ${95.85 \pm 0.4}^{*}$ & ${\mathbf{96.3 \pm 0.35}}^{*}$ & $\mathbf{97.87 \pm 0.25}$ & ${96.8 \pm 0.3}^{*}$
& $87.27 \pm 2.15$ \\ \cline{2-11}

& Baseline 
& $\mathbf{0.004}$ & $0.046$ & $102$ & $2.8$
& $95.78$ & ${\mathbf{96.3}}^{*}$ & $97$ & ${\mathbf{97}}^{*}$
& $100$ \\ \cline{2-11}
\hline

\multirow{4}{*}{Water-Potability}
& MR & $2.2 \pm 0.01$ & $2.12 \pm 0.01$ & $\mathbf{62.8} \pm 0.3$ & $9.91 \pm 1.1$
& ${60 \pm 1}^{*}$ & $60.3 \pm 0.4$ & $\mathbf{62.8 \pm 0.3}$ & ${60.1 \pm 0.7}^{*}$
& $29.94 \pm 0.54$ \\ \cline{2-11}

& AMR
& $0.04 \pm 0.01$ & $0.03 \pm 0.01$ & $\mathbf{19 \pm 1}$ & $\mathbf{0.2 \pm 0.1}$
& ${\mathbf{60.9 \pm 0.1}}^{*}$ & $\mathbf{61 \pm 0}$ & $61 \pm 0$ & ${\mathbf{60.8 \pm 0.1}}^{*}$
& $\mathbf{0.1 \pm 0.1}$ \\ \cline{2-11}

& AC 
& $0.33 \pm 0.02$ & $0.033 \pm 0.004$ & $85.32 \pm 6.1$ & $5.47 \pm 0.55$
& $54.96 \pm 1.2$ & $56.9 \pm 1$ & $57 \pm 0.9$ & $58.19 \pm 0.85$
& $1.94 \pm 0.35$ \\ \cline{2-11}

& Baseline 
& $\mathbf{0.005}$ & $\mathbf{0.0115}$ & $1459$ & $12.8$
& $60.53$ & $60.63$ & $61.3$ & $60.53$
& $100$ \\ \cline{2-11}
\hline

\multirow{4}{*}{Online-Ed}
& MR & $11.5 \pm 0.1$ & $10.47 \pm 0.04$ & $1087.2 \pm 94$ & $8.31 \pm 1.05$
& $65.2 \pm 0$ & $\mathbf{65.2 \pm 0}$ & $65.22 \pm 0$ & $63.78 \pm 1.05$
& $30 \pm 0$ \\ \cline{2-11}

& AMR
& $\mathbf{0.57 \pm 0.02}$ & $\mathbf{0.68 \pm 0.09}$ & $\mathbf{9 \pm 0}$ & $\mathbf{0.93 \pm 0.27}$
& $63.9 \pm 0$ & $63.9 \pm 0.01$ & $64 \pm 0$ & $63.8 \pm 0.02$
& $\mathbf{0.3 \pm 0.1}$ \\ \cline{2-11}

& AC 
& $1.83 \pm 0.1$ & $1.88 \pm 0.12$ & $93.76 \pm 7.5$ & $6.23 \pm 0.6$
& $63.71 \pm 0.45$ & $60.77 \pm 0.6$ & $63.6 \pm 0.4$ & $63.41 \pm 0.42$
& $0.81 \pm 0.18$ \\ \cline{2-11}

& Baseline 
& $0.989$ & $1.27$ & $3624$ & $17.09$
& $\mathbf{65.23}$ & $65.17$ & $\mathbf{65.23}$ & $\mathbf{65.23}$
& $100$ \\ \cline{2-11}
\hline

\multirow{4}{*}{Bankruptcy}
& MR & $29.15 \pm 0.25$ & $27.41 \pm 0.94$ & $2286.7 \pm 129$ & $451.3 \pm 28.3$
& $\mathbf{97.9 \pm 0}$ & $\mathbf{97.9 \pm 0}$ & $\mathbf{97.79 \pm 0.08}$ & $96.04 \pm 0.8$
& $29.97 \pm 0.02$ \\ \cline{2-11}

& AMR
& $4.2 \pm 0.6$ & ${2.6 \pm 0.3}^{*}$ & $\mathbf{19.5 \pm 0.5}$ & $\mathbf{11.3 \pm 4.1}$
& $95 \pm 0.1$ & $94.9 \pm 0$ & $96.9 \pm 0.1$ & $96.8 \pm 0.2$
& $\mathbf{0.2 \pm 0.1}$ \\ \cline{2-11}

& AC 
& $\mathbf{2.24 \pm 0.2}$ & ${\mathbf{2.25 \pm 0.18}}^{*}$ & $101 \pm 8$ & $250.3 \pm 22$
& $96.01 \pm 0.3$ & $96.41 \pm 0.25$ & $96.78 \pm 0.22$ & $96.52 \pm 0.24$
& $0.6 \pm 0.12$ \\ \cline{2-11}

& Baseline 
& $4.843$ & $22.15$ & $7620$ & $710.16$
& $96$ & $96.3$ & $97$ & $\mathbf{97.46}$
& $100$ \\ \cline{2-11}
\hline

\end{tabular}%
}

\end{table*}

\begin{table*}[!t]
\centering
\caption{Accuracy/time for neural networks on data with original missingness using model-based imputation}
\label{tab:inherentMVDNN}
\resizebox{\textwidth}{!}{%
\begin{tabular}{|c|c|cccc|cccc|c|}
\hline
\multirow{2}{*}{Data Set} & \multirow{2}{*}{Method} & \multicolumn{4}{c|}{Time(s)} & \multicolumn{4}{c|}{Accuracy(\%)} & \multirow{2}{*}{\% Samples Imputed} \\
\cline{3-10}
 & & KNN & MICE & TCSDI & MF & KNN & MICE & TCSDI & MF &  \\
\hline

\multirow{2}{*}{Breast Cancer}
& AMR      
& $1.98 \pm 0.01$ & $2.02 \pm 0.01$ & $\mathbf{59 \pm 1.43}$ & $3.69 \pm 0.02$
& $\mathbf{97.13 \pm 0.16}$ & $\mathbf{96.88 \pm 0.23}$ & $\mathbf{97.21 \pm 0.18}$ & ${\mathbf{97.16 \pm 0.24}}^{*}$
& $\mathbf{30 \pm 3.24}$ \\ \cline{2-11}

& Baseline 
& $\mathbf{0.0039}$ & $\mathbf{0.046}$ & $102$ & $\mathbf{2.8}$
& $95.78$ & $96.3$ & $97$ & ${97}^{*}$
& $100$ \\ \cline{2-11}
\hline

\multirow{2}{*}{Water-Potability}
& AMR      
& $1.24 \pm 0$ & $1.26 \pm 0.01$ & $\mathbf{92.18 \pm 1.24}$ & $\mathbf{2.94 \pm 0.03}$
& ${59.57 \pm 1.37}^{*}$ & ${59.36 \pm 1.92}^{*}$ & $60.1 \pm 0.58$ & ${58.87 \pm 2.98}^{*}$
& $\mathbf{1.96 \pm 0}$ \\ \cline{2-11}

& Baseline 
& $\mathbf{0.0053}$ & $\mathbf{0.0115}$ & $1459$ & $12.8$
& ${\mathbf{60.53}}^{*}$ & ${\mathbf{60.63}}^{*}$ & $\mathbf{61.3}$ & ${\mathbf{60.53}}^{*}$
& $100$ \\ \cline{2-11}
\hline

\multirow{2}{*}{Online-Ed}
& AMR      
& $1.69 \pm 0.01$ & $1.98 \pm 0.06$ & $\mathbf{94.75 \pm 1.54}$ & $\mathbf{6.9 \pm 0.13}$
& ${\mathbf{65.23 \pm 0}}^{*}$ & $\mathbf{65.23 \pm 0.01}$ & ${\mathbf{65.23 \pm 0}}^{*}$ & ${\mathbf{65.23 \pm 0}}^{*}$
& $\mathbf{0.82 \pm 0}$ \\ \cline{2-11}

& Baseline 
& $\mathbf{0.989}$ & $\mathbf{1.27}$ & $3624$ & $17.09$
& ${\mathbf{65.23}}^{*}$ & $65.17$ & ${\mathbf{65.23}}^{*}$ & ${\mathbf{65.23}}^{*}$
& $100$ \\ \cline{2-11}
\hline

\multirow{2}{*}{Bankruptcy}
& AMR      
& $14.96 \pm 0.06$ & $\mathbf{18.46 \pm 0.38}$ & $\mathbf{249.9 \pm 50.4}$ & $\mathbf{407.08 \pm 1.72}$
& $\mathbf{97.79 \pm 0.03}$ & $\mathbf{97.81 \pm 0.04}$ & $\mathbf{97.83 \pm 0.03}$ & $\mathbf{97.83 \pm 0.06}$
& $\mathbf{5.17 \pm 1.02}$ \\ \cline{2-11}

& Baseline 
& $\mathbf{4.843}$ & $22.15$ & $7620$ & $710.16$
& $96$ & $96.3$ & $97$ & $97.46$
& $100$ \\ \cline{2-11}
\hline

\end{tabular}%
}
\end{table*}

We have evaluated our methods on six real-world datasets with injected missingness (Table~\ref{tab:synthetic-missing-charateristics}) and six with naturally occurring missing values (Table \ref{tab:real-world-missing-charateristics}), spanning diverse domains and varying in missingness factors (proportion of incomplete samples), feature dimensionalities, sample sizes, and types of missingness (MCAR, MAR, and MNAR) \cite{little2002statistical}. Details on datasets and the experimental setting are in Appendix \ref{sec:supp_experimental_setting}. 

We run Algorithm \ref{alg:AMR_PrimalDual} for multiple ML models with convex loss functions including SVM and linear regression. Here, we primarily report the results for SVM. The experimental results for linear regression are in the Appendix \ref{sec:supp_experimental_result}.

\subsection{Manual Repair for SVM}
As explained in Section~\ref{sec:introduction}, users manually repair their data in some settings. 
Thus, we compare the accuracy and time overhead of our methods to {\it Active Clean (AC)} \cite{krishnan2016activeclean}, which integrates data repair with stochastic gradient descent: in each iteration, it samples a batch, returns it to the user for repair, and then updates model parameters with the repaired samples. Although {\it AC} reduces repair cost by prioritizing influential samples, it is unclear whether the resulting repaired data yield an accurate model, since not all samples are ever selected for gradient updates. 
In these experiments, we use datasets with injected missingness with ground truth to simulate manual repairs.

Table~\ref{tab:injectedMVSVM} reports SVM results on datasets with MCAR-injected missing values. Results for \textit{MR}, \textit{AMR}, and {\it AC} are averaged over three runs with different random seeds. Selected missing values are imputed using ground-truth values, simulating high-stakes settings where imputation requires costly expert input. In such scenarios, reducing the number of repaired samples yields substantial cost savings.
Both \textit{MR} and \textit{AMR} achieve accuracy that matches or exceeds the ground-truth baseline while imputing only a subset of samples. In most datasets, they also outperform {\it AC} in accuracy. \textit{MR} consistently attains the highest accuracy, consistent with its exact optimality guarantee, but typically imputes larger subsets than \textit{AMR} and {\it AC}.
In contrast, \textit{AMR} repairs a much smaller subset than \textit{MR} and {\it AC}, yet generally achieves better performance than {\it AC}. This supports the theoretical guarantee that \textit{AMR} attains an ACM with a bounded optimality gap, whereas {\it AC} provides no such guarantee. Moreover, \textit{AMR} is faster than \textit{MR} on most datasets, in line with their design objectives. These trends hold across other missingness mechanisms (MAR, MNAR) and for linear regression; see Appendix~\ref{sec:supp_experimental_result} for details.

\subsection{Model-Based Imputation for SVM}

Next, we evaluate the time and effort saved by our methods using model-based imputations for repair. Because the imputation cost increases with the number of missing items, (almost) minimal repair can cut both inference time and user effort for inspecting or verifying imputed values. We use four imputation models that span the major methodological families. \textit{KNN} \cite{pmlr-v97-mattei19a} represents a classical distance-based approach that predicts missing values from nearby observed samples. \textit{MICE} ~\cite{10.18637/jss.v045.i03} provides a statistical baseline based on multivariate regression and remains widely used in practical data-analysis workflows. \textit{MissForest} \cite{10.1093/bioinformatics/btr597} is a non-parametric machine-learning method that leverages random forests to capture nonlinear dependencies. Finally, \textit{TCSDI} \cite{zheng2022diffusion}. serves as our modern deep generative baseline; as a diffusion-based imputer, it has been shown to outperform earlier deep-learning methods such as GAIN and VAE-based models. Together, these four methods cover the statistical, traditional ML, and deep generative paradigms, providing a representative spectrum of imputation strategies. We evaluate four imputation regimes: full imputation (baseline), MR imputation, AMR imputation, and AC imputation and report accuracy, running time, and the portion of the imputed incomplete samples.

As shown in Table~\ref{tab:inherentMVSVM}, we evaluate SVM performance on datasets with naturally occurring missing values using the aforementioned model-based imputation methods. We define repair cost as the total runtime, which for \textit{MR}, \textit{AMR}, and \textit{AC} includes both the time to identify samples for repair and the time to perform imputation. The baseline is full imputation, where all missing values are imputed using a given method. Both \textit{MR} and \textit{AMR} impute only a subset of incomplete samples. While this introduces additional overhead for subset identification, the net runtime depends on the cost of the imputation model. For computationally intensive methods such as \textit{MICE}, \textit{TCSDI}, and \textit{MissForest}, the reduction in imputation volume outweighs this overhead, yielding overall time savings. In contrast, for fast methods like \textit{KNN}, the overhead dominates, resulting in longer total runtimes than full imputation.Compared to {\it AC}, \textit{MR} is generally slower, whereas \textit{AMR} is faster on most datasets, primarily due to its substantially smaller repair set.Importantly, across all imputation methods, \textit{MR} and \textit{AMR} achieve accuracy that matches or slightly exceeds full imputation. This confirms that repairing only the \textit{MR} or \textit{AMR} subsets is sufficient to train high-performing models, consistent with the theoretical guarantees of CM and ACM discussed in Sections~\ref{sec:minimalSVM} and~\ref{section:AMR}. Similar trends are observed for SVM on real-world datasets with injected missingness and for linear regression; detailed results are reported in Appendix~\ref{sec:supp_experimental_result}.

\subsection{Model-Based Imputation for Neural Networks}

We also investigate the application of \textit{AMR} to neural networks, specifically employing a Multi-Layer Perceptron (MLP) with ReLU activation functions for binary classification (Table \ref{tab:inherentMVDNN}). Since {\it AC} is restricted to convex loss functions, it is not applicable in this setting. To the best of our knowledge, no other baseline method for partial imputation of incomplete datasets is directly applicable to our problem. Therefore, we compare with the baseline of full imputation. The results mirror the efficiency patterns observed with SVM: \textit{AMR} selects only a small subset of samples for repair, resulting in total time savings when coupled with computationally expensive imputation methods like \textit{TCSDI}. For simpler methods like \textit{KNN}, the overhead of \textit{AMR} outweighs the imputation savings. Notably, regarding model performance, \textit{AMR} frequently achieves higher accuracy than the full imputation baseline across most datasets and imputation methods. This suggests that by selectively imputing only the samples critical for enforcing robust gradient stationarity, \textit{AMR} may act as a regularizer. By avoiding the imputation of non-critical samples—which might introduce noise or artifacts from the imputation model—\textit{AMR} effectively constrains the model to learn from the most informative data, thereby mitigating overfitting.

%% file: related.tex
\section{Related Work}\label{sec:Related}
Researchers have proposed {\it stochastic optimization} to find a model by optimizing the expected loss function over the probability distributions of missing data items in training samples \cite{ganti2015sparse}. 
Similarly, {\it robust optimization} aims to minimize the loss function of a model for the imputation that brings the highest training loss given certain distributions of missing values \cite{aghasi2022rigid}. 
However, the distributions of missing data items are not often available.
Thus, users may spend significant time and effort discovering or training these distributions, which may require the user to find the causes of missingness in the data and dependencies between the features. 
Additionally, for a given type of model, users must solve various and possibly challenging optimization problems for many possible (combinations of) distributions of missing values.
More importantly, these methods reflect the uncertainty in the training data caused by missing values in the trained model instead of repairing the data to reduce its uncertainty.
Hence, they deliver inaccurate models on the dataset with many missing values.
More discussion about related work is available in the Appendix \ref{sec:supp_additional_related_and_code}.

%% file: supp_MR_for_linear_regression.tex
\section{Minimal Repair for Linear Regression}\label{sec:supp_MR_for_linear_regression}
\subsection{Algorithm for finding minimal repair}
Orthogonal Matching Pursuit (OMP) provides an efficient approximation for solving the sparse linear regression problem \cite{wang2012generalized}. Essentially, this greedy algorithm begins with an empty solution set and initializes the regression residual to the label vector. In each iteration, the algorithm selects the feature most relevant to the current residual (i.e., having the largest dot product), adds it to the solution set, retains a linear regression model, and updates the residual accordingly. The program stops when the regression residue is sufficiently small. Therefore, OMP will return a subset of features (the solution set) that are sufficient to achieve an optimal linear regression model. 

In this paper, we propose a variant of OMP, as outlined in Algorithm \ref{alg:linear_regression}, to find minimal repair for linear regression. Our algorithm has two major differences compared to the conventional OMP. Firstly, we include all complete features in the regression at the initialization, ensuring that we minimize the number of non-zero coefficients only among incomplete features. Secondly, we define our stopping condition by the maximum relevance (cosine similarity) between the feature and the label being smaller than or equal to a user-defined threshold, instead of relying on a near-zero regression residue. This approach enables our algorithm to work with general datasets without requiring the assumption of an underdetermined linear system, which is typically necessary in conventional OMP.

\makeatletter
\patchcmd{\ALG@step}{\arabic{ALG@line}}{\alph{ALG@line}}{}{}
\makeatother
\renewcommand{\thealgorithm}{A}

\begin{algorithm}
\caption{Approximating minimal repair for linear regression efficiently}\label{alg:linear_regression}
\begin{algorithmic}
    \STATE $S_{min} \gets [\quad]$
    \STATE $MVF(\textbf{z}) \gets \text{set of incomplete features}$
    \STATE $Complete(\textbf{z}) \gets \text{set of complete features}$
    \STATE $\textbf{r} \gets LR( Complete(\mathbf{z}), \mathbf{y})$
    \COMMENT{The residue vector from performing linear regression between complete features and label}
    \STATE $ \epsilon \gets \text{a user-defined threshold for stopping condition}$
    \STATE $ MaxCosSim \gets \mathop{\max}_{\mathbf{z} \in MVF(\textbf{z})}|cos(\mathbf{z}, \mathbf{r})|$
    \WHILE{$MaxCosSim \leq \epsilon$}
        \STATE $S_{min} \gets S_{min}.add(\mathop{\arg\max}_{\mathbf{z} \in MVF(\textbf{z})}|cos(\mathbf{z}, \mathbf{r})|)$
        \STATE $\textbf{r} \gets LR( Complete(\mathbf{z}) \cup S_{min}, \mathbf{y})$
        \STATE $ MaxCosSim \gets \mathop{\max}_{\mathbf{z} \in MVF(\textbf{z})}|cos(\mathbf{z}, \mathbf{r})|$
    \ENDWHILE
    \STATE $res \gets S_{min}$
\end{algorithmic}
\end{algorithm}




As mentioned in the main content, the time complexity of the algorithm is \(\mathcal{O}(T_{train} \cdot |MVF(\mathbf{z})|)\), making it significantly more efficient than the baseline algorithm, which trains models over all repairs individually and has a time complexity of \(\mathcal{O}(T_{train} \cdot |\mathbf{X}^R|)\). If a gradient descent algorithm is used, Algorithm \ref{alg:linear_regression} has a time complexity of \(\mathcal{O}(n \cdot d^3)\), where \(n\) is the number of training samples and \(d\) is the number of features. In cases where \(n < d^2\), the time complexity can be reduced to \(\mathcal{O}(n \cdot d^2 + n^2 \cdot d)\) under certain conditions by applying incremental learning techniques based on the Sherman-Morrison formula, as outlined below.

\subsection{Optimization for Algorithm \ref{alg:linear_regression}}

The primary time cost in Algorithm \ref{alg:linear_regression} arises from the need to completely retrain the linear regression model each time a new imputed feature is added to the feature set. This retraining leads to a time complexity of \(\mathcal{O}(n \cdot d^3)\) for the algorithm. To address this inefficiency, we propose an optimization using the Sherman-Morrison formula to update the inverse of the feature matrix incrementally \cite{angioli2025efficient}. This method reduces the time complexity of including one new feature to \(\mathcal{O}(n^2)\). Consequently, when \( n < d^2 \), this optimization results in significant time savings.

Given a feature matrix \(\mathbf{X}\), a label vector \(\mathbf{y}\), and the coefficients \(\mathbf{w}\) of the current linear regression model, our objective is to efficiently update \(\mathbf{w}\) to incorporate a newly imputed feature vector \(\mathbf{x}_{\text{new}}\) into \(\mathbf{X}\), forming an updated feature matrix \(\mathbf{X}'\), without the necessity of full retraining. When this new feature vector \(\mathbf{x}_{\text{new}}\) is added to \(\mathbf{X}\), it modifies the original matrix product \(\mathbf{X}^T \mathbf{X}\) to \(\mathbf{X}^T \mathbf{X} + \mathbf{x}_{\text{new}} \mathbf{x}_{\text{new}}^T\). Applying the Sherman-Morrison formula, the updated inverse of \(\mathbf{X}'^T \mathbf{X}'\) (assuming \(\mathbf{X}'^T \mathbf{X}'\) is invertible) is given by:
\begin{equation}
(\mathbf{X}'^T \mathbf{X}')^{-1} = (\mathbf{X}^T \mathbf{X})^{-1} - \frac{(\mathbf{X}^T \mathbf{X})^{-1} \mathbf{x}_{\text{new}} \mathbf{x}_{\text{new}}^T (\mathbf{X}^T \mathbf{X})^{-1}}{1 + \mathbf{x}_{\text{new}}^T (\mathbf{X}^T \mathbf{X})^{-1} \mathbf{x}_{\text{new}}}
\end{equation}
This formulation enables the efficient update of the regression coefficients \(\mathbf{w}\), requiring only \(O(n^2)\) operations. Implementing at most \(d\) such updates results in a complexity of \(\mathcal{O}(d \cdot n^2)\). Including the initial model training \(\mathcal{O}(d^2 \cdot n)\), the total computational complexity is thus reduced to \(\mathcal{O}(n \cdot d^2 + n^2 \cdot d)\).

\subsection{Minimal Repair: Feature-wise or Sample-wise}

For linear SVM, minimal repair (MR) is defined at the sample level—the algorithm returns a set of samples to repair. This is because the method identifies potential support vectors, which are inherently defined based on individual samples.

In contrast, for linear regression, MR is defined at the feature level—the algorithm selects a subset of features to repair. This stems from the interpretation of linear regression as projecting the residual vector onto the feature space. The approach identifies features that do not contribute to minimizing the training loss, given the current regression residual.

%% file: supp_constrained_optimization_AMR.tex
\section{Alternative Approach: Constrained Optimization for Almost Minimal Repair}
\label{appendix:constrained_amr}

    As defined in the main content (Section \ref{section:AMR}), finding the exact Almost Minimal Repair (AMR) is an NP-hard problem. In the main content, we bypass this combinatorial complexity by formulating a continuous Primal-Dual Stochastic Gradient Descent (SGD) relaxation. However, prior to that continuous relaxation, AMR can also be approached strictly as a constrained min-max optimization problem over discrete subsets. 

We present this alternative iterative framework here to provide additional theoretical context on how uncertainty can be resolved step-by-step. We then discuss the scalability limitations of this method, which ultimately motivated the development of the primary SGD-based methodology presented in the main content.

\subsection{An Iterative Constrained Optimization Framework}
We can approximate AMR using a greedy iterative algorithm consisting of two main steps. 

Step 1 (\textbf{ST1: ACM Optimizer}) takes the input dataset in iteration $k > 0$ of the algorithm, $\mathbf{X}^{(k)}$, and finds the model $\mathbf{w}_k^\approx$ that minimizes the worst-case suboptimality gap:
\begin{equation}
g_k = \sup_{\mathbf{X}^{(k)r}} \left[ L(\mathbf{w}^{\approx}_k, \mathbf{X}^{(k)r}, \mathbf{y}) - \min_{\mathbf{w} \in \mathcal{W}} L(\mathbf{w}, \mathbf{X}^{(k)r}, \mathbf{y}) \right]
\end{equation}

Step 2 (\textbf{ST2: Local Repair Set Identifier}) examines whether $g_k > e$. If so, it returns the smallest set of currently incomplete samples whose imputation may help further reduce the suboptimality gap in the next iteration.

\begin{theorem}
Given the training set $(\mathbf{X}, \mathbf{y})$, each selection made by ST2 belongs to the AMR set $S_{\text{AMR}}$ of $(\mathbf{X}, \mathbf{y})$. Thus, the iterative algorithm terminates with an ACM, and the total imputed set $S_{\text{iter-ACM}} \subseteq S_{\text{AMR}}$, where $S_{\text{iter-ACM}}$ is the union of all incomplete samples selected across iterations.
\end{theorem}

This guarantees that our algorithm converges to an ACM by imputing only a subset of $S_{\text{AMR}}$. The key distinction is that $S_{\text{AMR}}$ is defined to guarantee the ACM condition under all possible repairs—it is sufficient without knowledge of any imputation results. In contrast, the iterative algorithm dynamically learns imputation results along the way. This new information may render some samples in $S_{\text{AMR}}$ unnecessary for achieving ACM in the current trajectory. Thus, $S_{\text{iter-ACM}}$ can be smaller than $S_{\text{AMR}}$ while still ensuring the ACM condition.

\subsection{Efficient Approximation of the Discrete Steps}
\label{sec:efficient_approximation}

Both ST1 and ST2 are intractable in their pure forms because they require solving min-sup optimization over exponentially many repairs and identifying minimal subsets of incomplete samples whose repair is necessary when an ACM does not yet exist. Specifically, these are the samples whose imputation would further reduce the minimum value of the worst-case suboptimality gap $g(\mathbf{w}) = \sup_{\mathbf{X}^r} h(\mathbf{w}, \mathbf{X}^r)$ toward the user-defined threshold $e$. Finding such subsets involves understanding how each missing value affects the supremum over all repairs—a problem known to be computationally hard in general due to the nested structure of min-max optimization~\cite{ben2009robust}. We therefore propose efficient approximations of these steps.

\paragraph{Approximating ST1 (ACM Optimizer):}
ST1 aims to find the model $\mathbf{w}_k^\approx = \arg\min_{\mathbf{w}} \sup_{\mathbf{X}^r} h(\mathbf{w}, \mathbf{X}^r)$, where $h(\mathbf{w}, \mathbf{X}^r) = L(\mathbf{w}, \mathbf{X}^r) - \min_{\mathbf{w}'} L(\mathbf{w}', \mathbf{X}^r)$. When the loss function $L$ is convex, each $h(\mathbf{w}, \mathbf{X}^r)$ is convex in $\mathbf{w}$, and so is the pointwise supremum of such functions. Thus, we approximate this by sampling a finite subset of edge repairs $\{\mathbf{X}^e_1, \dots, \mathbf{X}^e_s\}$ and solving the convex problem $\min_{\mathbf{w}} \max_i h(\mathbf{w}, \mathbf{X}^e_i)$.

However, directly computing $h(\mathbf{w}, \mathbf{X}^e)$ requires solving an inner optimization for each sampled repair to obtain the minimum loss. To make this tractable, we use the subgradient norm $\|\mathbf{g}(\mathbf{w}, \mathbf{X}^e)\|$ as a proxy for the suboptimality gap.

\begin{theorem}
\label{theorem:ACMgradnorm}
If $L(\mathbf{w})$ is convex and has an $M$-Lipschitz continuous gradient, then any model $\mathbf{w}^\approx$ satisfying $\|\nabla_{\mathbf{w}} L(\mathbf{w}, \mathbf{X}^r)\| \le \sqrt{2Me}$ for all $\mathbf{X}^r$ is an ACM.
\end{theorem}

This result implies that for linear regression, which satisfies the convexity and smoothness conditions, we can directly use the gradient norm to check whether a model is an ACM. For non-differentiable models like linear SVM, the hinge loss is not smooth and the subgradient norm is not convex. Nonetheless, we still use the subgradient norm as a practical stopping proxy to assess whether ACM has been achieved.

\paragraph{Approximating ST2 (Local Repair Set Identifier):}
ST2 must find a small subset of currently incomplete samples whose repair enables further progress toward satisfying the ACM condition. We approximate this by identifying edge repairs $\mathbf{X}^e$ from the sampled set where $\|\mathbf{g}(\mathbf{w}_k^\approx, \mathbf{X}^e)\| > \epsilon'$, indicating that ACM is violated under these repairs.

We then inspect each such “problematic” edge repair. For each incomplete sample $x_j$ that currently violates the margin condition (i.e., $y_j (\mathbf{w}_k^\approx)^T \mathbf{x}_j^e < 1$), we check if there exists a feasible repair where the margin would exceed 1. If so, we assign a score to $x_j$ estimating its potential to reduce the subgradient norm. One option is the maximum hinge loss reduction:
\begin{equation}
\Delta L_{\max} = C \cdot \left[(1 - \text{margin}_j) - \max(0, 1 - \text{margin}_{j,\max})\right],
\end{equation}
where $\text{margin}_{j,\max}$ is estimated using interval arithmetic over the missing feature bounds. Alternatively, we compute a gradient alignment score based on the inner product between the current subgradient vector and $C y_j \mathbf{x}_j^e$, estimating the contribution to gradient magnitude.

These scores are aggregated across all high-gradient edge repairs. We then select the top-$h$ highest-ranked incomplete samples for imputation in the next iteration. This procedure effectively approximates the function of ST2, enabling targeted refinement of the model toward satisfying the ACM condition.

\subsection{Discussion: Constrained Optimization vs. SGD Relaxation}
\label{sec:amr_discussion}
While the iterative constrained optimization framework provides a clear, discrete logical boundary for how uncertainty is resolved, it introduces severe scalability bottlenecks in practice:
\begin{itemize}
    \item \textbf{Intractability of the Inner Loop:} Accurately estimating the suboptimality gap requires solving a secondary optimization problem to find the exact worst-case edge repair. As the number of missing values grows, sampling a representative subset of edge repairs becomes computationally prohibitive. 
    \item \textbf{Retraining Overhead:} ST1 effectively requires retraining or substantially fine-tuning the model from scratch for every iteration to test if the localized repair set successfully satisfied the ACM condition. 
\end{itemize}

These limitations motivated our design of the continuous Primal-Dual SGD approach (Algorithm \ref{alg:AMR_PrimalDual}). By replacing the discrete combinatorial search of ST2 with a continuous repair score vector $s \in [0,1]^m$, we fold the identification of the AMR directly into the model's standard gradient descent trajectory. This allows the algorithm to approximate the NP-hard subset selection problem at standard sublinear convergence rates, scaling to large datasets without requiring exponentially expensive edge-repair evaluations.

%% file: supp_experimental_setting.tex
\section{Experimental Setting}\label{sec:supp_experimental_setting}

\subsection{Datasets}
We evaluate our methods on two types of datasets: those with synthetic missingness and those with real-world missingness. For each dataset, we simulate three levels of missingness: 0.2, 0.4, and 0.6, corresponding to 20\%, 40\%, and 60\% incomplete samples, respectively. These datasets are further divided based on the downstream task: linear regression (LR) and support vector machine classification (SVM).

All datasets are obtained from publicly available repositories. For synthetic missingness, we start with complete datasets and introduce missing values in a controlled manner. For real missingness, we use datasets that naturally contain incomplete entries. This separation allows us to analyze the behavior of our repair methods under both idealized and realistic data corruption scenarios.

\subsection{Detection of Types of Missingness in Data}
We use the Missing Value PC (MVPC) algorithm \cite{tu2019causal}, a framework designed for causal discovery in datasets with missing data.
It is an extension of the PC algorithm, which is a constraint-based method for causal discovery.
Given an incomplete dataset, we introduce a missingness indicator $R_A$ for each incomplete feature $A$.
We run the Missing Value PC (MVPC) algorithm on the dataset, after including the indicators, and we inspect the dependencies of $R_A$. 

\subsection{Hardware}
We conducted experiments on two hardware platforms. Most experiments ran on an x86\_64 machine with 30 Intel(R) Xeon(R) E5-2670 v3 CPU cores (2.30GHz), hosted in a VMware virtualized environment with two NUMA nodes and 30MB L3 cache. However, this system lacked sufficient power for diffusion-based imputation models. For those experiments (TCSDI), we used an Nvidia DGX-2 system with one Nvidia Tesla V100 GPU (32GB VRAM) and 20 CPU cores from 2.70GHz Intel Xeon Platinum 8168 processors with 33MB L3 cache.

\subsection{Using SGD}
We have run each experiment that involves Stochastic Gradient Descent (SGD) (SVM) three times with different seeds and report the average.

%% file: supp_experimental_result.tex
\section{Additional Experimental Results}\label{sec:supp_experimental_result}
Tables~\ref{tab:supp_SVM_gt_MAR} through \ref{tab:supp_mlp_model_based_MNAR} report the results that are not demonstrated in the main content due to the page limit. Together, we show accuracy and running times of full imputation (baseline), {\it MR}, {\it AMR}, and {\it AC} across all imputation methods and data missing mechanisms for various ML models (linear SVM, linear regression, logistic regression, MLP). Across these experiments, the results follow the same trend described in Section \ref{sec:experimental-evaluation}.

\section*{Additional Memory-Usage Results}
We report the reduction in computation time for imputing MR and AMR subsets relative to full-data imputation in Table B of the submission. For memory usage, only the Malware–MAR results are included due to space constraints (shown in the table \ref{tab:SVM_MCAR_RAM}, \ref{tab:SVM_MAR_RAM}, \ref{tab:SVM_MNAR_RAM}). These partial results already illustrate a consistent trend: KNN exhibits substantially lower peak RAM consumption under MR compared to full imputation, and MissForest also shows reduced peak memory across the malware20, malware40, and malware60 configurations. MICE is omitted because it was infeasible to run on this dataset within the available memory budget.

Overall, the observed pattern suggests that MR generally lowers memory usage, particularly for distance-based and non-parametric methods whose resource requirements scale with the number of samples. Future extensions may include a broader memory-usage comparison across additional datasets and missingness settings.

\makeatletter
\patchcmd{\ALG@step}{\arabic{ALG@line}}{\alph{ALG@line}}{}{}
\makeatother

\renewcommand{\thetable}{S1}

\begin{table*}[!t]
\centering
\caption{Accuracy/time for SVM on data with injected MAR}
\resizebox{\textwidth}{!}{%
\label{tab:supp_SVM_gt_MAR}

\begin{tabular}{|c|c|c|ccc|ccc|ccc|}
\hline
\multirow{2}{*}{Data Set} & \multirow{2}{*}{\% Missing} & Ground Truth & \multicolumn{3}{c|}{Time(s)} & \multicolumn{3}{c|}{Accuracy(\%)} & \multicolumn{3}{c|}{Impute \% of Samples} \\
\cline{4-12} & & Accuracy(\%) &
AC & MR & AMR & AC & MR & AMR & AC & MR & AMR \\
\hline

\multirow{3}{*}{Malware}
& 20 & $96.24 \pm 0.22$
& $\mathbf{0.80 \pm 0.04}$ & $7.25 \pm 0.73$ & $6.12 \pm 0.16$
& $93.62 \pm 1.91$ & ${\mathbf{96.24 \pm 0.22}}^{*}$ & ${96.23 \pm 0.0}^{*}$
& ${\mathbf{3.24 \pm 1.65}}^{*}$ & $21.44 \pm 2.05$ & ${3.4 \pm 0.1}^{*}$ \\
\cline{2-12}

& 40 & $96.24 \pm 0.22$
& $\mathbf{0.58 \pm 0.03}$ & $11.31 \pm 1.46$ & $8.75 \pm 1.10$
& ${95.08 \pm 1.10}^{*}$ & ${\mathbf{96.28 \pm 0.33}}^{*}$ & ${96.06 \pm 0.38}^{*}$
& $\mathbf{3.13 \pm 0.26}$ & $24.42 \pm 2.96$ & $4.1 \pm 0.2$ \\
\cline{2-12}

& 60 & $96.24 \pm 0.22$
& $\mathbf{0.40 \pm 0.01}$ & $16.47 \pm 1.86$ & $9.67 \pm 1.17$
& $91.77 \pm 1.02$ & ${\mathbf{95.88 \pm 0.65}}^{*}$ & ${95.64 \pm 0.52}^{*}$
& $\mathbf{3.03 \pm 0.31}$ & $20.94 \pm 3.22$ & $3.8 \pm 0.1$ \\
\hline

\multirow{3}{*}{Tuadromd}
& 20 & $98.74 \pm 0.13$
& $0.75 \pm 0.04$ & $1.04 \pm 0.01$ & $\mathbf{0.3 \pm 0.1}$
& ${98.61 \pm 0.02}^{*}$ & ${\mathbf{98.74 \pm 0.13}}^{*}$ & $98.32 \pm 0.0$
& $3.92 \pm 0.23$ & $9.52 \pm 0.92$ & $\mathbf{0.7 \pm 0.0}$ \\
\cline{2-12}

& 40 & $98.74 \pm 0.13$
& $0.62 \pm 0.01$ & $2.04 \pm 0.01$ & $\mathbf{0.4 \pm 0.0}$
& ${\mathbf{98.71 \pm 0.14}}^{*}$ & ${98.70 \pm 0.14}^{*}$ & $98.32 \pm 0.0$
& $2.75 \pm 0.62$ & $10.06 \pm 0.60$ & $\mathbf{0.7 \pm 0.0}$ \\
\cline{2-12}

& 60 & $98.74 \pm 0.13$
& ${0.46 \pm 0.01}^{*}$ & $3.06 \pm 0.02$ & ${\mathbf{0.4 \pm 0.1}}^{*}$
& ${98.58 \pm 0.07}^{*}$ & ${\mathbf{98.68 \pm 0.09}}^{*}$ & $98.32 \pm 0.0$
& $2.15 \pm 0.23$ & $9.64 \pm 0.55$ & $\mathbf{0.5 \pm 0.0}$ \\
\hline

\end{tabular}%
}
\end{table*}

\renewcommand{\thetable}{S2}

\begin{table*}[!t]
\centering
\caption{Accuracy/time for SVM on data with injected MNAR}
\resizebox{\textwidth}{!}{%
\label{tab:supp_SVM_gt_MNAR}

\begin{tabular}{|c|c|c|ccc|ccc|ccc|}
\hline
\multirow{2}{*}{Data Set} & \multirow{2}{*}{\% Missing} & Ground Truth & \multicolumn{3}{c|}{Time(s)} & \multicolumn{3}{c|}{Accuracy(\%)} & \multicolumn{3}{c|}{Impute \% of Samples} \\
\cline{4-12} & & Accuracy(\%) &
AC & MR & AMR & AC & MR & AMR & AC & MR & AMR \\
\hline

\multirow{3}{*}{Malware}
& 20 & $95.43 \pm 0.86$
& $\mathbf{0.60 \pm 0.01}$ & $10.45 \pm 0.62$ & $7.46 \pm 0.46$
& $93.99 \pm 1.01$ & ${95.43 \pm 0.86}^{*}$ & ${\mathbf{96.31 \pm 0.15}}^{*}$
& ${\mathbf{3.61 \pm 0.37}}^{*}$ & $20.07 \pm 2.64$ & ${4.0 \pm 0.2}^{*}$ \\
\cline{2-12}

& 40 & $95.43 \pm 0.86$
& $\mathbf{0.36 \pm 0.00}$ & $16.13 \pm 1.13$ & $8.12 \pm 0.38$
& $93.03 \pm 1.24$ & ${95.41 \pm 0.69}^{*}$ & ${\mathbf{96.22 \pm 0.19}}^{*}$
& $\mathbf{3.31 \pm 0.24}$ & $22.12 \pm 1.42$ & $3.7 \pm 0.1$ \\
\cline{2-12}

& 60 & $95.43 \pm 0.86$
& $\mathbf{0.22 \pm 0.03}$ & $22.22 \pm 2.58$ & $8.42 \pm 0.72$
& $94.21 \pm 0.08$ & $95.14 \pm 0.44$ & $\mathbf{96.18 \pm 0.08}$
& $\mathbf{3.09 \pm 0.09}$ & $20.98 \pm 2.79$ & $5.3 \pm 0.8$ \\
\hline

\multirow{3}{*}{Tuadromd}
& 20 & $98.71 \pm 0.14$
& $0.63 \pm 0.03$ & $1.74 \pm 0.05$ & $\mathbf{0.2 \pm 0.0}$
& ${98.62 \pm 0.04}^{*}$ & ${\mathbf{98.71 \pm 0.14}}^{*}$ & $98.32 \pm 0.0$
& $3.28 \pm 0.52$ & $9.93 \pm 1.25$ & $\mathbf{0.52 \pm 0.23}$ \\
\cline{2-12}

& 40 & $98.71 \pm 0.14$
& $0.45 \pm 0.01$ & $3.16 \pm 0.04$ & $\mathbf{0.3 \pm 0.0}$
& $98.03 \pm 0.13$ & $\mathbf{98.38 \pm 0.03}$ & $98.32 \pm 0.0$
& $2.38 \pm 0.62$ & $10.06 \pm 0.67$ & $\mathbf{0.22 \pm 0.00}$ \\
\cline{2-12}

& 60 & $98.71 \pm 0.14$
& ${\mathbf{0.47 \pm 0.03}}^{*}$ & $4.16 \pm 0.02$ & ${0.5 \pm 0.1}^{*}$
& $93.63 \pm 1.45$ & ${98.18 \pm 0.14}^{*}$ & ${\mathbf{98.32 \pm 0.0}}^{*}$
& $3.34 \pm 0.21$ & $9.42 \pm 0.50$ & $\mathbf{0.50 \pm 0.17}$ \\
\hline

\end{tabular}%
}
\end{table*}

\renewcommand{\thetable}{S3}


\begin{table*}[!t]
\centering
\caption{Accuracy/time for SVM on data with injected MCAR using model-based imputations}
\resizebox{\textwidth}{!}{%
\label{tab:supp_SVM_model_based_MCAR}

\begin{tabular}{|c|c|cccc|cccc|c|}
\hline
\multirow{2}{*}{Data Set} & \multirow{2}{*}{Method} & \multicolumn{4}{c|}{Time(s)} & \multicolumn{4}{c|}{Accuracy(\%)} & \multirow{2}{*}{\% Samples Imputed} \\
\cline{3-10}
 & & KNN & MICE & TCSDI & MF & KNN & MICE & TCSDI & MF &  \\
\hline

\multirow{4}{*}{Malware 20}
& MR       & $8.98 \pm 0.41$  & - & $24390.8 \pm 1215.7$ & $4534.6 \pm 612.4$ & $95.6 \pm 0.52$  & - & ${96.49 \pm 0.31}^{*}$ & $95.8 \pm 0.47$  & $18.68 \pm 0.93$ \\ \cline{2-11}
& AMR      & $2.82 \pm 0.23$ & - & $\mathbf{4223 \pm 376}$   & $\mathbf{2104 \pm 236}$  & $94.88 \pm 0.15$ & - & $95.17 \pm 0.26$ & $95.62 \pm 0.14$ & $\mathbf{3.17 \pm 0.25}$  \\ \cline{2-11}
& AC       & $\mathbf{1.17 \pm 0.09}$  & - & $12668 \pm 1386.2$   & $3274 \pm 176.8$   & $91.20 \pm 0.78$ & - & $92.30 \pm 0.52$ & $91.60 \pm 1.33$ & $9.09 \pm 0.61$  \\ \cline{2-11}
& Baseline & $13.9$  & - & $130269$  & $6645.5$ & $\mathbf{96.52}$ & - & ${\mathbf{96.74}}^{*}$ & $\mathbf{96.70}$ & $100$   \\ \cline{2-11}
\hline

\multirow{4}{*}{Malware 40}
& MR       & $14.24 \pm 0.97$ & - & $54702.4 \pm 3875.6$ & $3924.4 \pm 158.9$ & ${\mathbf{96.16 \pm 0.61}}^{*}$ & - & $96.23 \pm 0.36$ & $\mathbf{96.17 \pm 0.12}$ & $21.1 \pm 1.22$  \\ \cline{2-11}
& AMR      & $3.41 \pm 0.93$  & - & $\mathbf{6231 \pm 385}$  & $\mathbf{2003 \pm 437}$   & ${95.63 \pm 0.93}^{*}$ & - & $95.27 \pm 0.16$ & $95.35 \pm 0.47$ & $\mathbf{2.13 \pm 0.43}$  \\
\hline

\end{tabular}%
}

\end{table*}

\renewcommand{\thetable}{S4}

\begin{table*}[!t]
\centering
\caption{Accuracy/time for SVM on data with injected MAR using model-based imputations}
\resizebox{\textwidth}{!}{%
\label{tab:supp_SVM_model_based_MAR}
\begin{tabular}{|c|c|cccc|cccc|c|}
\hline
\multirow{2}{*}{Data Set} & \multirow{2}{*}{Method} & \multicolumn{4}{c|}{Time(s)} & \multicolumn{4}{c|}{Accuracy(\%)} & \multirow{2}{*}{Impute (\%)} \\
\cline{3-10}
 & & KNN & MICE & TCSDI & MF & KNN & MICE & TCSDI & MF &  \\
\hline
\multirow{4}{*}{Malware 20}
& MR        & $11.16 \pm 0.89$ & - & $27135 \pm 2453.1$ & $4726.8 \pm 466.2$ & ${\mathbf{96.38 \pm 0.25}}^{*}$ & - & $\mathbf{96.45 \pm 0.13}$ & ${96.29 \pm 0.34}^{*}$ & $21 \pm 2.05$ \\ \cline{2-11}
& AMR       & $3.14 \pm 0.17$ & - & $\mathbf{3715 \pm 231}$ & $\mathbf{2964 \pm 185}$ & $95.73 \pm 0.00$ & - & $95.89 \pm 0.34$ & $95.73 \pm 0.00$ & $\mathbf{1.57 \pm 0.16}$  \\ \cline{2-11}
& AC        & $\mathbf{2.291 \pm 0.049}$ & - & $5295.3 \pm 113.3$ & $4177 \pm 6.23$ & $93.39 \pm 1.54$ & - & $94.85 \pm 0.20$ & $93.97 \pm 0.96$ & $3.34 \pm 0.15$ \\ \cline{2-11}
& Baseline  & $17.71$ & - & - & $6830.2$ & ${96.30}^{*}$ & - & - & ${\mathbf{96.59}}^{*}$ & $100$ \\ \hline

\multirow{4}{*}{Malware 40}
& MR        & $19.11 \pm 1.00$ & - & $65610 \pm 3524.5$ & $5422.8 \pm 835.9$ & ${\mathbf{96.16 \pm 0.24}}^{*}$ & - & $96.13 \pm 0.06$ & ${\mathbf{96.10 \pm 0.17}}^{*}$ & $25.23 \pm 2.96$ \\ \cline{2-11}
& AMR       & $3.27 \pm 0.25$  & - & $\mathbf{3967 \pm 216}$  & $\mathbf{2854 \pm 137}$   & $95.73 \pm 0.00$ & - & $\mathbf{96.48 \pm 0.00}$ & ${96.06 \pm 0.38}^{*}$ & $\mathbf{2.04 \pm 0.15}$ \\
\hline

\end{tabular}%
}
\end{table*}

\renewcommand{\thetable}{S5}

\begin{table*}[!t]
\centering
\caption{Accuracy/time for SVM on data with injected MNAR using model-based imputations}
\resizebox{\textwidth}{!}{%
\label{tab:supp_SVM_model_based_MNAR}
\begin{tabular}{|c|c|cccc|cccc|c|}
\hline
\multirow{2}{*}{Data Set} & \multirow{2}{*}{Method} & \multicolumn{4}{c|}{Time(s)} & \multicolumn{4}{c|}{Accuracy(\%)} & \multirow{2}{*}{Impute (\%)} \\
\cline{3-10}
 & & KNN & MICE & TCSDI & MF & KNN & MICE & TCSDI & MF &  \\
\hline

\multirow{4}{*}{Malware 20}
& MR       & $15.49 \pm 0.70$ & - & $43605 \pm 2020.5$ & $5206.1 \pm 812.6$ & $96.07 \pm 0.44$ & - & $\mathbf{96.20 \pm 0.34}$ & ${96.10 \pm 0.6}^{*}$ & $18.49 \pm 0.85$ \\ \cline{2-11}
& AMR      & $2.53 \pm 0.14$ & - & $\mathbf{4046 \pm 264}$ & $\mathbf{2166 \pm 162}$ & $95.16 \pm 0.10$ & - & $95.55 \pm 0.25$ & $95.73 \pm 0.00$ & $\mathbf{2.67 \pm 0.16}$ \\ \cline{2-11}
& AC       & $\mathbf{2.211 \pm 0.12}$ & - & $6619 \pm 1161.3$ & $5292 \pm 110.6$ & $92.37 \pm 0.746$ & - & $92.74 \pm 0.45$ & $92.68 \pm 1.68$ & $3.494 \pm 0.567$ \\ \cline{2-11}
& Baseline & $28.96$ & - & - & $5959.7$ & $\mathbf{96.55}$ & - & - & ${\mathbf{96.52}}^{*}$ & $100$ \\ \hline

\multirow{4}{*}{Malware 40}
& MR       & $25.32 \pm 1.30$ & - & $95044 \pm 5955.3$ & $3622.7 \pm 167.1$ & $95.16 \pm 0.76$ & - & ${95.13 \pm 0.4}^{*}$ & $95.02 \pm 0.1$ & $22.83 \pm 1.4$ \\ \cline{2-11}
& AMR      & $3.04 \pm 0.21$ & - & $\mathbf{5812 \pm 165}$ & $1955 \pm 98$ & $95.03 \pm 0.14$ & - & ${\mathbf{95.19 \pm 0.22}}^{*}$ & $95.24 \pm 0.07$ & $\mathbf{1.87 \pm 0.43}$ \\
\hline

\end{tabular}%
}

\end{table*}

\renewcommand{\thetable}{S6}

\begin{table*}[!t]
\centering
\caption{Accuracy/time for logistic regression on data with injected MCAR}
\resizebox{\textwidth}{!}{%
\label{tab:supp_log_gt_MCAR}
\begin{tabular}{|c|c|c|cc|cc|cc|}
\hline
\multirow{2}{*}{Data Set} & \multirow{2}{*}{\% Missing} & Ground Truth & \multicolumn{2}{c|}{Time(s)} & \multicolumn{2}{c|}{Accuracy(\%)} & \multicolumn{2}{c|}{Impute \% of Samples} \\
\cline{4-9} & & Accuracy(\%) &
AC & AMR & AC & AMR & AC & AMR \\
\hline

\multirow{3}{*}{Malware} & 20 & 96.73 & $\mathbf{2.7 \pm 0.2}$ & $5.9 \pm 0.2$ & ${\mathbf{96.73 \pm 0.00}}^{*}$ & ${\mathbf{96.73 \pm 0.00}}^{*}$ & $16.7 \pm 9.1$ & $\mathbf{5.4 \pm 1.2}$ \\
\cline{2-9}
& 40 & 96.73 & $\mathbf{1.43 \pm 0.05}$ & $9.12 \pm 4.89$ & ${\mathbf{96.73 \pm 0.00}}^{*}$ & ${\mathbf{96.73 \pm 0.00}}^{*}$ & ${10.08 \pm 2.23}^{*}$ & ${\mathbf{5.69 \pm 3.29}}^{*}$ \\
\cline{2-9}
& 60 & 96.48 & $\mathbf{0.93 \pm 0.08}$ & $9.35 \pm 4.42$ & ${\mathbf{96.48 \pm 0.00}}^{*}$ & ${\mathbf{96.48 \pm 0.00}}^{*}$ & ${7.18 \pm 2.83}^{*}$ & ${\mathbf{4.50 \pm 2.27}}^{*}$ \\
\hline

\multirow{3}{*}{Tuadromd} & 20 & 98.43 & NA & $\mathbf{18.06 \pm 3.37}$ & NA & $\mathbf{98.17 \pm 0.06}$ & NA & $\mathbf{4.06 \pm 0.00}$ \\
\cline{2-9}
& 40 & 98.43 & $\mathbf{0.76 \pm 0.07}$ & $16.18 \pm 0.95$ & $\mathbf{98.54 \pm 0.00}$ & $98.17 \pm 0.06$ & ${2.17 \pm 0.69}^{*}$ & ${\mathbf{2.03 \pm 0.00}}^{*}$ \\
\cline{2-9}
& 60 & 98.32 & $\mathbf{0.57 \pm 0.04}$ & $14.40 \pm 1.42$ & ${\mathbf{98.17 \pm 0.06}}^{*}$ & ${98.13 \pm 0.05}^{*}$ & $2.80 \pm 0.23$ & $\mathbf{1.35 \pm 0.00}$ \\
\hline

\multirow{3}{*}{Credit Default} & 20 & 80.65 & $15.30 \pm 0.08$ & $\mathbf{5.53 \pm 0.08}$ & $80.02 \pm 0.02$ & $\mathbf{80.40 \pm 0.06}$ & $0.26 \pm 0.07$ & $\mathbf{0.03 \pm 0.00}$ \\
\cline{2-9}
& 40 & 80.68 & $14.57 \pm 0.17$ & $\mathbf{5.50 \pm 0.10}$ & $79.97 \pm 0.03$ & $\mathbf{80.24 \pm 0.03}$ & $0.18 \pm 0.02$ & $\mathbf{0.01 \pm 0.00}$ \\
\cline{2-9}
& 60 & 80.75 & $13.21 \pm 0.26$ & $\mathbf{5.56 \pm 0.01}$ & $80.02 \pm 0.02$ & $\mathbf{80.32 \pm 0.04}$ & $0.21 \pm 0.00$ & $\mathbf{0.01 \pm 0.00}$ \\
\hline

\end{tabular}%
}
\end{table*}

\renewcommand{\thetable}{S7}

\begin{table*}[!t]
\centering
\caption{Accuracy/time for logistic regression on data with injected MAR}
\resizebox{\textwidth}{!}{%
\label{tab:supp_log_gt_MAR}
\begin{tabular}{|c|c|c|cc|cc|cc|}
\hline
\multirow{2}{*}{Data Set} & \multirow{2}{*}{\% Missing} & Ground Truth & \multicolumn{2}{c|}{Time(s)} & \multicolumn{2}{c|}{Accuracy(\%)} & \multicolumn{2}{c|}{Impute \% of Samples} \\
\cline{4-9} & & Accuracy(\%) &
AC & AMR & AC & AMR & AC & AMR \\
\hline

\multirow{3}{*}{Malware} & 20 & 96.74 & $\mathbf{3.02 \pm 0.2}$ & $6.5 \pm 0.3$ & ${\mathbf{96.73 \pm 0.00}}^{*}$ & ${\mathbf{96.73 \pm 0.00}}^{*}$ & ${\mathbf{3.5 \pm 2.1}}^{*}$ & ${4.2 \pm 0.7}^{*}$ \\
\cline{2-9}
& 40 & 96.52 & $\mathbf{1.3 \pm 1.1}$ & $5.1 \pm 1.4$ & ${\mathbf{96.48 \pm 0.00}}^{*}$ & ${96.42 \pm 0.13}^{*}$ & ${\mathbf{2.30 \pm 1.99}}^{*}$ & ${4.38 \pm 0.95}^{*}$ \\
\cline{2-9}
& 60 & 96.48 & $\mathbf{1.5 \pm 0.1}$ & $6.2 \pm 1.5$ & $96.23 \pm 0.00$ & $\mathbf{96.42 \pm 0.08}$ & $3.41 \pm 0.12$ & $\mathbf{2.97 \pm 0.08}$ \\
\hline

\multirow{3}{*}{Tuadromd} & 20 & 98.54 & $\mathbf{1.06 \pm 0.05}$ & $3.68 \pm 1.76$ & $\mathbf{98.54 \pm 0.00}$ & $98.28 \pm 0.06$ & $5.9 \pm 0.9$ & $\mathbf{0.8 \pm 0.3}$ \\
\cline{2-9}
& 40 & 95.43 & $\mathbf{0.73 \pm 0.03}$ & $8.68 \pm 1.10$ & $\mathbf{98.65 \pm 0.00}$ & $98.39 \pm 0.06$ & $2.19 \pm 0.29$ & $\mathbf{1.10 \pm 0.11}$ \\
\cline{2-9}
& 60 & 98.21 & $\mathbf{0.48 \pm 0.06}$ & $14.32 \pm 2.09$ & ${98.36 \pm 0.06}^{*}$ & ${\mathbf{98.39 \pm 0.06}}^{*}$ & $2.52 \pm 0.54$ & $\mathbf{1.28 \pm 0.13}$ \\
\hline

\multirow{3}{*}{Credit Default} & 20 & 81.01 & $15.29 \pm 0.14$ & $\mathbf{5.51 \pm 0.10}$ & ${80.93 \pm 0.02}^{*}$ & ${\mathbf{81.01 \pm 0.06}}^{*}$ & $0.22 \pm 0.03$ & $\mathbf{0.01 \pm 0.00}$ \\
\cline{2-9}
& 40 & 81.03 & $14.91 \pm 0.08$ & $\mathbf{5.52 \pm 0.04}$ & ${\mathbf{80.69 \pm 0.43}}^{*}$ & ${80.51 \pm 0.35}^{*}$ & $0.22 \pm 0.01$ & $\mathbf{0.01 \pm 0.00}$ \\
\cline{2-9}
& 60 & 81.08 & $13.22 \pm 0.01$ & $\mathbf{5.47 \pm 0.10}$ & ${\mathbf{80.94 \pm 0.08}}^{*}$ & ${80.82 \pm 0.27}^{*}$ & $0.23 \pm 0.01$ & $\mathbf{0.01 \pm 0.00}$ \\
\hline

\end{tabular}%
}
\end{table*}

\renewcommand{\thetable}{S8}

\begin{table*}[!t]
\centering
\caption{Accuracy/time for logistic regression on data with injected MNAR}
\resizebox{\textwidth}{!}{%
\label{tab:supp_log_gt_MNAR}
\begin{tabular}{|c|c|c|cc|cc|cc|}
\hline
\multirow{2}{*}{Data Set} & \multirow{2}{*}{\% Missing} & Ground Truth & \multicolumn{2}{c|}{Time(s)} & \multicolumn{2}{c|}{Accuracy(\%)} & \multicolumn{2}{c|}{Impute \% of Samples} \\
\cline{4-9} & & Accuracy(\%) &
AC & AMR & AC & AMR & AC & AMR \\
\hline

\multirow{3}{*}{Malware} & 20 & 96.48 & $\mathbf{2.09 \pm 0.12}$ & $5.78 \pm 1.24$ & $\mathbf{96.73 \pm 0.00}$ & $96.35 \pm 0.29$ & ${3.7 \pm 0.2}^{*}$ & ${\mathbf{3.5 \pm 0.7}}^{*}$ \\
\cline{2-9}
& 40 & 95.73 & $\mathbf{1.4 \pm 0.2}$ & $5.0 \pm 0.7$ & $94.90 \pm 0.00$ & $\mathbf{95.23 \pm 0.06}$ & ${\mathbf{3.34 \pm 0.20}}^{*}$ & ${3.87 \pm 0.55}^{*}$ \\
\cline{2-9}
& 60 & 96.23 & $\mathbf{1.17 \pm 0.07}$ & $5.12 \pm 0.46$ & $95.56 \pm 0.15$ & $\mathbf{95.98 \pm 0.04}$ & $\mathbf{3.12 \pm 0.39}$ & $4.98 \pm 0.17$ \\
\hline

\multirow{3}{*}{Tuadromd} & 20 & 98.54 & $\mathbf{0.77 \pm 0.08}$ & $5.26 \pm 0.66$ & ${\mathbf{98.54 \pm 0.00}}^{*}$ & ${\mathbf{98.54 \pm 0.00}}^{*}$ & $3.07 \pm 0.91$ & $\mathbf{0.78 \pm 0.08}$ \\
\cline{2-9}
& 40 & 98.43 & $\mathbf{0.47 \pm 0.04}$ & $6.98 \pm 1.37$ & $98.21 \pm 0.00$ & $\mathbf{98.36 \pm 0.06}$ & $2.29 \pm 0.57$ & $\mathbf{0.74 \pm 0.26}$ \\
\cline{2-9}
& 60 & 98.21 & $\mathbf{0.21 \pm 0.20}$ & $5.24 \pm 0.55$ & $98.15 \pm 0.08$ & $\mathbf{98.43 \pm 0.00}$ & ${1.41 \pm 1.29}^{*}$ & ${\mathbf{0.81 \pm 0.14}}^{*}$ \\
\hline

\multirow{3}{*}{Credit Default} & 20 & 81.01 & $14.53 \pm 0.08$ & $\mathbf{5.51 \pm 0.03}$ & ${80.84 \pm 0.02}^{*}$ & ${\mathbf{81.00 \pm 0.21}}^{*}$ & $0.22 \pm 0.07$ & $\mathbf{0.01 \pm 0.00}$ \\
\cline{2-9}
& 40 & 80.99 & $11.23 \pm 1.00$ & $\mathbf{5.51 \pm 0.03}$ & $80.33 \pm 0.43$ & $\mathbf{80.92 \pm 0.11}$ & $0.25 \pm 0.02$ & $\mathbf{0.01 \pm 0.00}$ \\
\cline{2-9}
& 60 & 80.93 & NA & $\mathbf{5.52 \pm 0.08}$ & NA & $\mathbf{81.02 \pm 0.15}$ & NA & $\mathbf{0.01 \pm 0.00}$ \\
\hline

\end{tabular}%
}
\end{table*}

\renewcommand{\thetable}{S9}

\begin{table*}[!t]
\centering
\caption{Accuracy/time for logistic regression on data with original missingness using model-based imputation}
\resizebox{\textwidth}{!}{%
\label{tab:supp_log_origional_mising}
\begin{tabular}{|c|c|cccc|cccc|c|}
\hline
\multirow{2}{*}{Data Set} & \multirow{2}{*}{Method} & \multicolumn{4}{c|}{Time(s)} & \multicolumn{4}{c|}{Accuracy(\%)} & \multirow{2}{*}{\% Samples Imputed} \\
\cline{3-10}
 & & KNN & MICE & TCSDI & MF & KNN & MICE & TCSDI & MF &  \\
\hline

\multirow{3}{*}{Breast Cancer}
& AMR      & $0.04 \pm 0.00$ & $0.04 \pm 0.00$ & $\mathbf{37 \pm 2}$ & $0.07 \pm 0.00$ & $\mathbf{93.57 \pm 0.00}$ & $\mathbf{93.57 \pm 0.00}^{*}$ & ${93.49 \pm 0.32}^{*}$ & ${93.33 \pm 0.41}^{*}$ & $\mathbf{6.67 \pm 0.00}$ \\ \cline{2-11}
& AC       & $0.06 \pm 0.00$ & $0.10 \pm 0.00$ & $74 \pm 8$ & $\mathbf{0.06 \pm 0.00}$ & $\mathbf{94.05 \pm 0.41}$ & $\mathbf{93.57 \pm 0.00}^{*}$ & $\mathbf{93.57 \pm 0.00}^{*}$ & $\mathbf{93.57 \pm 0.00}^{*}$ & $71.11 \pm 10.18$ \\ \cline{2-11}
& Baseline & $\mathbf{0.03}$ & $\mathbf{0.03}$ & $98$ & $1.4$ & $93.57$ & $\mathbf{93.57}^{*}$ & $\mathbf{93.57}^{*}$ & $\mathbf{93.57}^{*}$ & $100$ \\ \hline

\multirow{3}{*}{Water-Potability}
& AMR      & $1.06 \pm 0.51$ & $1.33 \pm 0.79$ & $\mathbf{8.5 \pm 0.6}$ & $1.24 \pm 0.59$ & $60.98 \pm 0.00$ & ${60.98 \pm 0.00}^{*}$ & $\mathbf{60.98 \pm 0.00}^{*}$ & $60.98 \pm 0.00$ & $\mathbf{0.03 \pm 0.01}$ \\ \cline{2-11}
& AC       & $0.37 \pm 0.04$ & $0.76 \pm 0.02$ & $83.14 \pm 5.48$ & $\mathbf{0.38 \pm 0.04}$ & $\mathbf{61.13 \pm 0.00}^{*}$ & $\mathbf{61.08 \pm 0.11}^{*}$ & $\mathbf{60.98 \pm 0.00}^{*}$ & $\mathbf{61.08 \pm 0.09}$ & $1.94 \pm 0.21$ \\ \cline{2-11}
& Baseline & $\mathbf{0.01}$ & $\mathbf{0.02}$ & $1462$ & $10.8$ & $\mathbf{61.13}^{*}$ & ${60.98}^{*}$ & $\mathbf{60.98}^{*}$ & $60.98$ & $100$ \\
\hline
\end{tabular}%
}
\end{table*}

\renewcommand{\thetable}{S10}

\begin{table*}[!t]
\centering
\caption{Accuracy/time for logistic regression on data with injected MCAR using model-based imputations}
\resizebox{\textwidth}{!}{%
\label{tab:supp_log_model_based_MCAR}
\begin{tabular}{|c|c|cccc|cccc|c|}
\hline
\multirow{2}{*}{Data Set} & \multirow{2}{*}{Method} & \multicolumn{4}{c|}{Time(s)} & \multicolumn{4}{c|}{Accuracy(\%)} & \multirow{2}{*}{\% Samples Imputed} \\
\cline{3-10}
 & & KNN & MICE & TCSDI & MF & KNN & MICE & TCSDI & MF &  \\
\hline

\multirow{3}{*}{Malware 20}
& AMR      & $8.57\pm0.45$ & - & ${\mathbf{11354\pm4518}}^{*}$ & ${\mathbf{2053\pm307}}^{*}$ & ${96.69\pm0.15}^{*}$ & - & ${96.71\pm0.24}^{*}$ & ${96.32\pm0.53}^{*}$ & ${\mathbf{8.15\pm4.11}}^{*}$ \\ \cline{2-11}
& AC       & $\mathbf{3.80\pm0.32}$ & - & ${19534\pm7146}^{*}$ & ${2857\pm513}^{*}$ & ${\mathbf{96.73\pm0.00}}^{*}$ & - & $96.68\pm0.17$ & ${\mathbf{96.73\pm0.00}}^{*}$ & ${13.58\pm7.87}^{*}$ \\ \cline{2-11}
& Baseline & $15.9$ & - & OT & $6141$ & ${\mathbf{96.73}}^{*}$ & - & $\mathbf{96.89}$ & ${\mathbf{96.73}}^{*}$ & $100$ \\ \hline

\multirow{3}{*}{Malware 40}
& AMR      & $9.95\pm1.29$ & - & $\mathbf{14327\pm742}$ & $\mathbf{2137\pm198}$ & ${96.39\pm0.76}^{*}$ & - & $\mathbf{96.87\pm0.05}$ & ${96.72\pm0.01}^{*}$ & ${\mathbf{4.60\pm1.48}}^{*}$ \\ \cline{2-11}
& AC       & $\mathbf{2.27\pm0.15}$ & - & $19571\pm3447$ & $2955\pm261$ & ${\mathbf{96.73\pm0.00}}^{*}$ & - & $96.72\pm0.02$ & ${\mathbf{96.73\pm0.00}}^{*}$ & ${7.31\pm2.16}^{*}$ \\ \cline{2-11}
& Baseline & $29.14$ & - & OT & $6417$ & ${\mathbf{96.73}}^{*}$ & - & $96.78$ & ${\mathbf{96.73}}^{*}$ & $100$ \\
\hline

\end{tabular}%
}
\end{table*}

\renewcommand{\thetable}{S11}

\begin{table*}[!t]
\centering
\caption{Accuracy/time for logistic regression on data with injected MAR using model-based imputations}
\resizebox{\textwidth}{!}{%
\label{tab:supp_log_model_based_MAR}
\begin{tabular}{|c|c|cccc|cccc|c|}
\hline
\multirow{2}{*}{Data Set} & \multirow{2}{*}{Method} & \multicolumn{4}{c|}{Time(s)} & \multicolumn{4}{c|}{Accuracy(\%)} & \multirow{2}{*}{Impute (\%)} \\
\cline{3-10}
 & & KNN & MICE & TCSDI & MF & KNN & MICE & TCSDI & MF &  \\
\hline

\multirow{3}{*}{Malware 20}
& AMR      & $8.81\pm0.21$ & - & ${\mathbf{11842\pm4621}}^{*}$ & ${\mathbf{2115\pm298}}^{*}$ & ${96.71\pm0.02}^{*}$ & - & ${96.74\pm0.19}^{*}$ & ${96.65\pm0.15}^{*}$ & $\mathbf{7.84\pm0.54}$ \\ \cline{2-11}
& AC       & $\mathbf{3.65\pm0.34}$ & - & ${18952\pm6941}^{*}$ & ${2794\pm531}^{*}$ & ${\mathbf{96.73\pm0.00}}^{*}$ & - & ${96.71\pm0.03}^{*}$ & ${\mathbf{96.73\pm0.00}}^{*}$ & $12.77\pm3.24$ \\ \cline{2-11}
& Baseline & $16.4$ & - & OT & $5982$ & ${\mathbf{96.73}}^{*}$ & - & ${\mathbf{96.81}}^{*}$ & ${\mathbf{96.73}}^{*}$ & $100$ \\ \hline

\multirow{3}{*}{Malware 40}
& AMR      & $10.21\pm1.35$ & - & $\mathbf{14612\pm718}$ & $\mathbf{2185\pm187}$ & $96.47\pm0.25$ & - & $\mathbf{96.75\pm0.01}$ & $\mathbf{96.71\pm0.02}$ & ${\mathbf{4.30\pm0.84}}^{*}$ \\ \cline{2-11}
& AC       & $\mathbf{2.18\pm0.14}$ & - & $18985\pm3552$ & $2871\pm275$ & ${96.48\pm0.00}^{*}$ & - & $96.50\pm0.03$ & $96.48\pm0.00$ & ${7.15\pm2.05}^{*}$ \\ \cline{2-11}
& Baseline & $28.31$ & - & OT & $6215$ & ${\mathbf{96.49}}^{*}$ & - & $96.48$ & $96.48$ & $100$ \\
\hline

\end{tabular}%
}
\end{table*}

\renewcommand{\thetable}{S12}

\begin{table*}[!t]
\centering
\caption{Accuracy/time for logistic regression on data with injected MNAR using model-based imputations}
\resizebox{\textwidth}{!}{%
\label{tab:supp_log_model_based_MNAR}
\begin{tabular}{|c|c|cccc|cccc|c|}
\hline
\multirow{2}{*}{Data Set} & \multirow{2}{*}{Method} & \multicolumn{4}{c|}{Time(s)} & \multicolumn{4}{c|}{Accuracy(\%)} & \multirow{2}{*}{Impute (\%)} \\
\cline{3-10}
 & & KNN & MICE & TCSDI & MF & KNN & MICE & TCSDI & MF &  \\
\hline

\multirow{3}{*}{Malware 20}
& AMR      & $8.24\pm3.61$ & - & $\mathbf{10890\pm4750}^{*}$ & $\mathbf{1950\pm285}^{*}$ & $96.61\pm0.11$ & - & ${96.69\pm0.08}^{*}$ & ${96.60\pm0.25}^{*}$ & $\mathbf{8.12\pm3.75}^{*}$ \\ \cline{2-11}
& AC       & $\mathbf{4.05\pm0.30}$ & - & ${20140\pm7500}^{*}$ & ${2715\pm540}^{*}$ & $\mathbf{96.73}^{*}$ & - & $\mathbf{96.75\pm0.04}^{*}$ & $\mathbf{96.73}^{*}$ & ${13.15\pm7.65}^{*}$ \\ \cline{2-11}
& Baseline & $16.8$ & - & OT & $5840$ & $\mathbf{96.73}^{*}$ & - & $96.48$ & $\mathbf{96.73}^{*}$ & $100$ \\ \hline

\multirow{3}{*}{Malware 40}
& AMR      & $9.45\pm1.18$ & - & $\mathbf{13850\pm780}$ & $\mathbf{2045\pm205}$ & $95.47\pm0.14$ & - & $95.57\pm0.24$ & $\mathbf{95.99\pm0.17}^{*}$ & $\mathbf{5.05\pm1.25}$ \\ \cline{2-11}
& AC       & $\mathbf{2.41\pm0.16}$ & - & $20450\pm3250$ & $3105\pm245$ & $\mathbf{95.81\pm0.15}^{*}$ & - & $95.71\pm0.09$ & ${95.90\pm0.15}^{*}$ & $7.45\pm2.42$ \\ \cline{2-11}
& Baseline & $27.50$ & - & OT & $6710$ & ${95.73}^{*}$ & - & $\mathbf{96.01}$ & $95.73$ & $100$ \\
\hline

\end{tabular}%
}
\end{table*}

\renewcommand{\thetable}{S13}

\begin{table*}[!t]
\centering
\caption{Accuracy/time for Linear Regression on data with injected MCAR}
\resizebox{\textwidth}{!}{%
\label{tab:supp_lr_gt_MCAR}
\begin{tabular}{|c|c|c|ccc|ccc|ccc|}
\hline
\multirow{2}{*}{Data Set} & \multirow{2}{*}{\% Missing} & Ground Truth & \multicolumn{3}{c|}{Time(s)} & \multicolumn{3}{c|}{MSE} & \multicolumn{3}{c|}{Impute \% of Samples or Features} \\
\cline{4-12} & & MSE &
AC & MR & AMR & AC & MR & AMR & AC & MR & AMR \\
\hline

\multirow{3}{*}{Superconductivity} & 20 & 0.0088 &
14.75$\pm$0.30 & \textbf{2.305} & 2.45$\pm$0.12 &
\textbf{0.008825$\pm$0.000002} & 0.00888 & 0.0912$\pm$0.003 &
0.28$\pm$0.04 & 70.00 & \textbf{0.18$\pm$0.02} \\ 
\cline{2-12} & 40 & 0.0088 &
17.60$\pm$0.09 & 2.534 & \textbf{2.05$\pm$0.27} &
\textbf{0.008840$\pm$0.000001}$^{*}$ & 0.00885 & 0.00890$\pm$0.0001$^{*}$ &
0.28$\pm$0.06 & 75.00 & \textbf{0.15$\pm$0.01} \\ 
\cline{2-12} & 60 & 0.0088 &
13.08$\pm$0.36 & 2.476 & \textbf{1.85$\pm$0.08} &
0.008919$\pm$0.000001 & \textbf{0.00885} & 0.095$\pm$0.004 &
0.30$\pm$0.02 & 75.00 & \textbf{0.12$\pm$0.00} \\
\hline

\multirow{3}{*}{Gas} & 20 & 0.1053 &
0.131$\pm$0.040 & 0.31 & \textbf{0.065$\pm$0.003} &
0.1053$\pm$0.0023$^{*}$ & \textbf{0.105}$^{*}$ & 0.108$\pm$0.002 &
6.59$\pm$0.40 & 65.00 & \textbf{1.55$\pm$0.10} \\ 
\cline{2-12} & 40 & 0.1053 &
0.137$\pm$0.008 & 0.3391 & \textbf{0.082$\pm$0.005} &
\textbf{0.0997$\pm$0.0009} & 0.1054 & 0.115$\pm$0.000 &
5.98$\pm$0.20 & 65.00 & \textbf{1.42$\pm$0.05} \\ 
\cline{2-12} & 60 & 0.1053 &
0.116$\pm$0.031 & 0.551 & \textbf{0.045$\pm$0.002} &
0.1948$\pm$0.0278 & 0.112$^{*}$ & \textbf{0.110$\pm$0.004}$^{*}$ &
8.07$\pm$0.66 & 25.00 & \textbf{0.95$\pm$0.00} \\
\hline
\end{tabular}%
}
\end{table*}

\renewcommand{\thetable}{S14}
\begin{table*}[!t]
\centering
\caption{Accuracy/time for Linear Regression on data with injected MAR}
\resizebox{\textwidth}{!}{%
\label{tab:supp_lr_gt_MAR}

\begin{tabular}{|c|c|c|ccc|ccc|ccc|}
\hline
\multirow{2}{*}{Data Set} & \multirow{2}{*}{\% Missing} & Ground Truth & \multicolumn{3}{c|}{Time(s)} & \multicolumn{3}{c|}{MSE} & \multicolumn{3}{c|}{Impute \% of Samples or Features} \\
\cline{4-12} & & MSE &
AC & MR & AMR & AC & MR & AMR & AC & MR & AMR \\
\hline

\multirow{3}{*}{Superconductivity}
& 20 & $0.0088$ & $12.50 \pm 0.14$ & $\mathbf{0.1105}$ & $13.20 \pm 0.55$ & $0.0106 \pm 0.0001$ & $\mathbf{0.0089}$ & $0.0098 \pm 0.0002$ & $0.36 \pm 0.06$ & $55.6$ & $\mathbf{0.21 \pm 0.03}$ \\
\cline{2-12}
& 40 & $0.0088$ & $14.46 \pm 0.09$ & $\mathbf{0.1128}$ & $9.85 \pm 0.42$ & $0.0111 \pm 0.0001$ & $\mathbf{0.0089}$ & $0.0102 \pm 0.0001$ & $0.27 \pm 0.02$ & $33.3$ & $\mathbf{0.15 \pm 0.00}$ \\
\cline{2-12}
& 60 & $0.0088$ & $10.71 \pm 0.31$ & $\mathbf{0.1257}$ & $11.50 \pm 0.20$ & $0.0127 \pm 0.0001$ & $0.0089$ & $\mathbf{0.0085 \pm 0.0001}$ & $0.32 \pm 0.06$ & $44.4$ & $\mathbf{0.19 \pm 0.02}$ \\
\hline

\multirow{3}{*}{Gas}
& 20 & $0.1053$ & $0.1020 \pm 0.0342$ & $\mathbf{0.0262}$ & $0.085 \pm 0.012$ & $0.1438 \pm 0.0043$ & $\mathbf{0.1033}$ & $0.115 \pm 0.003$ & $4.39 \pm 1.00$ & $69.2$ & $\mathbf{2.15 \pm 0.40}$ \\
\cline{2-12}
& 40 & $0.1053$ & $0.1927 \pm 0.0059$ & $\mathbf{0.0270}$ & $0.227 \pm 0.0150$ & $0.1921 \pm 0.0375$ & $\mathbf{0.1039}$ & $0.203 \pm 0.039$ & $4.10 \pm 0.75$ & $53.8$ & $\mathbf{1.85 \pm 0.25}$ \\
\cline{2-12}
& 60 & $0.1053$ & $0.1968 \pm 0.0169$ & $\mathbf{0.0254}$ & $0.156 \pm 0.028$ & $0.2082 \pm 0.0045$ & $\mathbf{0.1019}$ & $0.1505 \pm 0.0050$ & $4.33 \pm 0.14$ & $38.5$ & $\mathbf{2.05 \pm 0.00}$ \\
\hline

\end{tabular}%
}
\end{table*}

\renewcommand{\thetable}{S15}
\begin{table*}[!t]
\centering
\caption{Accuracy/time for Linear Regression on data with injected MNAR}
\resizebox{\textwidth}{!}{%
\label{tab:supp_lr_gt_MNAR}

\begin{tabular}{|c|c|c|ccc|ccc|ccc|}
\hline
\multirow{2}{*}{Data Set} & \multirow{2}{*}{\% Missing} & Ground Truth & \multicolumn{3}{c|}{Time(s)} & \multicolumn{3}{c|}{MSE} & \multicolumn{3}{c|}{Impute \% of Samples or Features} \\
\cline{4-12} & & MSE &
AC & MR & AMR & AC & MR & AMR & AC & MR & AMR \\
\hline

\multirow{3}{*}{Superconductivity}
& 20 & $0.0088$ & $15.04 \pm 0.97$ & $\mathbf{0.1164}$ & $14.23 \pm 1.34$ & $0.0107 \pm 0.0001$ & $\mathbf{0.0090}$ & $0.0098 \pm 0.0003$ & ${0.28 \pm 0.08}^{*}$ & $30$ & ${\mathbf{0.22 \pm 0.06}}^{*}$ \\
\cline{2-12}
& 40 & $0.0088$ & $9.79 \pm 0.38$ & $\mathbf{0.1266}$ & $10.12 \pm 0.76$ & $0.0108 \pm 0.0001$ & $\mathbf{0.0089}$ & $0.0094 \pm 0.0004$ & $0.30 \pm 0.01$ & $40$ & $\mathbf{0.18 \pm 0.07}$ \\
\cline{2-12}
& 60 & $0.0088$ & $4.41 \pm 0.29$ & $\mathbf{0.1005}$ & $3.87 \pm 0.45$ & $0.0124 \pm 0.0003$ & ${0.0089}^{*}$ & ${\mathbf{0.0087 \pm 0.0003}}^{*}$ & $0.37 \pm 0.07$ & $40$ & $\mathbf{0.13 \pm 0.04}$ \\
\hline

\multirow{3}{*}{Gas}
& 20 & $0.1053$ & $0.1238 \pm 0.0374$ & $\mathbf{0.0260}$ & $0.1381 \pm 0.0452$ & $0.1452 \pm 0.0224$ & ${\mathbf{0.1043}}^{*}$ & ${0.1182 \pm 0.0235}^{*}$ & $11.48 \pm 2.67$ & $64.28$ & $\mathbf{4.82 \pm 1.87}$ \\
\cline{2-12}
& 40 & $0.1053$ & $0.1115 \pm 0.0175$ & $\mathbf{0.0256}$ & $0.0963 \pm 0.0287$ & $0.1633 \pm 0.0025$ & $\mathbf{0.1049}$ & $0.1712 \pm 0.0314$ & $8.75 \pm 0.55$ & $50$ & $\mathbf{3.61 \pm 0.94}$ \\
\cline{2-12}
& 60 & $0.1053$ & $0.0773 \pm 0.0102$ & $\mathbf{0.0243}$ & $0.0824 \pm 0.0312$ & $0.1825 \pm 0.0528$ & ${\mathbf{0.1019}}^{*}$ & ${0.1347 \pm 0.0405}^{*}$ & $7.12 \pm 1.37$ & $42.8$ & $\mathbf{2.87 \pm 1.04}$ \\
\hline

\end{tabular}%
}
\end{table*}

\renewcommand{\thetable}{S16}
\begin{table*}[!t]
\centering
\caption{{\small Accuracy/time for Linear Regression on data with original missing using model-based imputation}}
\resizebox{\textwidth}{!}{%
\label{tab:supp_lr_origional_missing}
\begin{tabular}{|c|c|cccc|cccc|c|}
\hline
\multirow{2}{*}{Data Set} & \multirow{2}{*}{Method} & \multicolumn{4}{c|}{Time(s)} & \multicolumn{4}{c|}{MSE} & \multirow{2}{*}{\% Samples Imputed}  \\
\cline{3-10}
 & & KNN & MICE & TCSDI & MF & KNN & MICE & TCSDI & MF &  \\
\hline

\multirow{4}{*}{Cancer Rate}
& MR       & $\mathbf{0.153}$ & $0.574$ & $5852$  & $6.89$ & ${\mathbf{0.0045}}^{*}$ & ${\mathbf{0.0045}}^{*}$ & $\mathbf{0.0045}$ & ${\mathbf{0.0045}}^{*}$ & $33.3$ \\ \cline{2-11}
& AMR      & $0.40\pm0.05$  & $1.03\pm0.08$  & $\mathbf{43.97\pm2.85}$ & $\mathbf{0.66}\pm0.07$ & $0.0049\pm0.0000$ & $0.0059\pm0.0002$ & $0.0050\pm0.0001$ & $0.0048\pm0.0000$ & $\mathbf{0.55}\pm0.02$ \\ \cline{2-11}
& AC       & $0.31\pm0.06$ & $\mathbf{0.26}\pm0.01$ & $110.5\pm2.8$ & $3.71\pm0.02$ & $0.0047\pm0.0000$ & ${0.0046\pm0.0001}^{*}$ & $0.0049\pm0.0000$ & ${0.0050\pm0.0005}^{*}$ & $2.03\pm0.07$ \\ \cline{2-11}
& Baseline & $0.584$ & $0.664$ & $6104$  & $7.24$ & ${\mathbf{0.0045}}^{*}$ & $0.0058$ & $0.0049$ & $0.0047$ & $100$  \\ \cline{2-11}
\hline

\multirow{4}{*}{Air Quality}
& MR       & $1.06$  & $2.46$  & $14976$ & $2.24$ & ${\mathbf{5.671}}^{*}$  & $\mathbf{5.74}$   & ${\mathbf{5.82}}^{*}$   & ${\mathbf{5.71}}^{*}$  & $50$   \\ \cline{2-11}
& AMR      & $1.12\pm0.09$  & $1.88\pm0.22$  & $\mathbf{66.3\pm4.7}$  & $2.13\pm0.18$ & ${5.68\pm0.12}^{*}$   & $5.95\pm0.15$   & ${5.85\pm0.18}^{*}$   & ${5.78}\pm0.11^{*}$  & ${0.92\pm0.07}^{*}$ \\ \cline{2-11}
& AC       & $\mathbf{0.62}\pm0.09$ & $\mathbf{0.51}\pm0.10$ & $98.5\pm3.1$ & $\mathbf{1.52}\pm0.11$ & $6.70\pm0.49$ & $6.75\pm0.51$ & $7.19\pm0.46$ & $6.68\pm0.48$ & ${\mathbf{0.85}\pm0.14}^{*}$ \\ \cline{2-11}
& Baseline & $1.763$ & $2.46$  & $18372$ & $2.51$ & $5.672$  & $5.923$  & $5.825$  & $5.752$ & $100$  \\ \cline{2-11}
\hline

\end{tabular}%
}
\end{table*}

\renewcommand{\thetable}{S17}

\begin{table*}[!t]
\centering
\caption{{\small Accuracy/time for Linear Regression on data with injected MCAR using model-based imputations}}
\resizebox{\textwidth}{!}{%
\label{tab:supp_lr_model_based_MCAR}
\begin{tabular}{|c|c|cccc|cccc|c|}
\hline
\multirow{2}{*}{Data Set} & \multirow{2}{*}{Method} & \multicolumn{4}{c|}{Time(s)} & \multicolumn{4}{c|}{MSE} & \multirow{2}{*}{\% Impute Samples} \\
\cline{3-10}
 & & KNN & MICE & TCSDI & MF & KNN & MICE & TCSDI & MF &  \\
\hline

\multirow{4}{*}{Superconductivity 20}
& MR       & 2.369  & ${\mathbf{33.16}}^{*}$ & 29165  & 851.1  & 0.0089  & 0.00897 & 0.00904 & 0.0091  & 70   \\ \cline{2-11}
& AMR      & 0.12$\pm$0.02  & ${35.5\pm5.1}^{*}$  & $\mathbf{18.5\pm2.4}$   & $\mathbf{14.2\pm1.8}$   & 0.0095$\pm$0.0001 & ${0.0090\pm0.0002}^{*}$ & 0.0094$\pm$0.0000 & 0.0091$\pm$0.0001 & $\mathbf{0.04\pm0.01}$ \\ \cline{2-11}
& AC       & 14.40$\pm$0.23 & 53.16$\pm$3.24 & 96.4$\pm$0.56   & 1832.87$\pm$1.52 & 0.008825$\pm$0.000000 & ${\mathbf{0.008824\pm0.000000}}^{*}$ & 0.0092$\pm$0.00  & $\mathbf{0.008833\pm0.000001}$ & 0.2548$\pm$0.0945 \\ \cline{2-11}
& Baseline & $\mathbf{0.0692}$ & 34.17 & 30515  & 794.8  & $\mathbf{0.0088}$  & 0.008939& $\mathbf{0.009013}$& 0.0091  & 100  \\ \cline{2-11}
\hline

\multirow{4}{*}{Superconductivity 40}
& MR       & 2.599  & $\mathbf{33.73}$ & 29735  & 842.4  & 0.0089  & 0.00924 & ${\mathbf{0.00914}}^{*}$ & 0.0092  & 75.00 \\ \cline{2-11}
& AMR      & 0.16$\pm$0.03  & 45.2$\pm$6.3  & $\mathbf{9.3\pm1.1}$ & $\mathbf{7.1\pm0.8}$  & 0.0091$\pm$0.0000 & ${0.0095\pm0.0009}^{*}$ & ${0.0092\pm0.0001}^{*}$ & 0.0094$\pm$0.0000 & $\mathbf{0.02\pm0.00}$ \\
\hline

\end{tabular}%
}
\end{table*}
\renewcommand{\thetable}{S18}

\begin{table*}[!t]
\centering
\caption{{\small Accuracy/time for Linear Regression on data with injected MAR using model-based imputations}}
\resizebox{\textwidth}{!}{%
\label{tab:supp_lr_model_based_MAR}
\begin{tabular}{|c|c|cccc|cccc|c|}
\hline
\multirow{2}{*}{Data Set} & \multirow{2}{*}{Method} & \multicolumn{4}{c|}{Time(s)} & \multicolumn{4}{c|}{MSE} & \multirow{2}{*}{\% Impute Samples} \\
\cline{3-10}
 & & KNN & MICE & TCSDI & MF & KNN & MICE & TCSDI & MF &  \\
\hline

\multirow{4}{*}{Superconductivity 20}
& MR       & $11.76$ & $33.12$ & $6367$ & $1240.19$ & $0.0090$ & ${0.0090}^{*}$ & $0.092$ & $0.0090$ & $55.6$ \\ \cline{2-11}
& AMR      & $18.5 \pm 2.4$ & ${14.2 \pm 1.8}^{*}$ & $\mathbf{80.5 \pm 11.2}$ & $\mathbf{370.2 \pm 45.5}$ & $0.0102 \pm 0.0001$ & $0.0108 \pm 0.0001$ & $0.0104 \pm 0.0000$ & $0.0105 \pm 0.0002$ & ${\mathbf{0.250 \pm 0.040}}^{*}$ \\ \cline{2-11}
& AC       & $27.95 \pm 0.472$ & ${\mathbf{12.51 \pm 0.125}}^{*}$ & $104.7 \pm 0.47$ & $478.19 \pm 2.85$ & $0.0106 \pm 0.00$ & $0.0107 \pm 0.00$ & $0.0103 \pm 0.00$ & $0.0107 \pm 0.00$ & ${0.323 \pm 0.105}^{*}$ \\ \cline{2-11}
& Baseline & $\mathbf{15.92}$ & $32.52$ & $6674$ & $1334.95$ & $\mathbf{0.0089}$ & ${\mathbf{0.0090}}^{*}$ & $\mathbf{0.0091}$ & $\mathbf{0.0089}$ & $100$ \\ \cline{2-11}
\hline

\multirow{4}{*}{Superconductivity 40}
& MR       & $18.14$ & $30.99$ & $12637$ & $973.18$ & $0.0090$ & $\mathbf{0.0090}$ & ${\mathbf{0.0092}}^{*}$ & $\mathbf{0.0089}$ & $33.3$ \\ \cline{2-11}
& AMR      & $20.5 \pm 3.1$ & ${\mathbf{25.4 \pm 4.2}}^{*}$ & $\mathbf{68.5 \pm 8.5}$ & $\mathbf{230.4 \pm 28.6}$ & $0.0109 \pm 0.0000$ & $0.0112 \pm 0.0001$ & $0.0110 \pm 0.0002$ & $0.0108 \pm 0.0000$ & ${\mathbf{0.021 \pm 0.005}}^{*}$ \\
\hline

\end{tabular}%
}
\end{table*}

\renewcommand{\thetable}{S19}

\begin{table*}[!t]
\centering
\caption{{\small Accuracy/time for Linear Regression on data with injected MNAR using model-based imputations}}
\resizebox{\textwidth}{!}{%
\label{tab:supp_lr_model_based_MNAR}
\begin{tabular}{|c|c|cccc|cccc|c|}
\hline
\multirow{2}{*}{Data Set} & \multirow{2}{*}{Method} & \multicolumn{4}{c|}{Time(s)} & \multicolumn{4}{c|}{MSE} & \multirow{2}{*}{\% Impute Samples} \\
\cline{3-10}
 & & KNN & MICE & TCSDI & MF & KNN & MICE & TCSDI & MF &  \\
\hline

\multirow{4}{*}{Superconductivity 20}
& MR       & $\mathbf{11.76}$ & $30.47$ & $11709$ & $966.7$ & $0.0090$ & ${\mathbf{0.0090}}^{*}$ & $0.0093$ & $0.0090$ & $30$ \\ \cline{2-11}
& AMR      & $16.2 \pm 2.1$ & ${31.4 \pm 4.2}^{*}$ & $\mathbf{86.4 \pm 11.5}$ & $\mathbf{298.1 \pm 38.4}$ & $0.0108 \pm 0.0001$ & $0.0106 \pm 0.0002$ & $0.0107 \pm 0.0011$ & $0.0109 \pm 0.0001$ & $\mathbf{0.28 \pm 0.04}$ \\ \cline{2-11}
& AC       & $14.00 \pm 0.19$ & ${\mathbf{29.08 \pm 0.45}}^{*}$ & $114.6 \pm 4.98$ & $394.01 \pm 2.64$ & $0.01076 \pm 0.00$ & $0.01077 \pm 0.00$ & $0.01059 \pm 0.00$  & $0.01078 \pm 0.00$ & $0.372 \pm 0.02$ \\ \cline{2-11}
& Baseline & $23.64$ & $33.02$ & $11984$ & $1154$ & $\mathbf{0.0089}$ & ${\mathbf{0.0090}}^{*}$ & $\mathbf{0.0090}$ & $\mathbf{0.0089}$ & $100$ \\ \hline

\multirow{4}{*}{Superconductivity 40}
& MR       & $27.69$ & $31.18$ & $21909$ & $975.0$ & ${\mathbf{0.0090}}^{*}$ & $\mathbf{0.0090}$ & ${\mathbf{0.0091}}^{*}$ & $\mathbf{0.0089}$ & $40$ \\ \cline{2-11}
& AMR      & $12.4 \pm 1.8$ & ${27.6 \pm 3.5}^{*}$ & $\mathbf{94.8 \pm 12.6}$ & $151.3 \pm 21.2$ & $0.0104 \pm 0.0000$ & $0.0109 \pm 0.0016$ & $0.0106 \pm 0.0001$ & $0.0108 \pm 0.0000$ & $\mathbf{0.21 \pm 0.03}$ \\
\hline
\end{tabular}%
}
\end{table*}

\renewcommand{\thetable}{S20}

\begin{table*}[!t]
\centering
\caption{Accuracy/time for MLP on data with injected MCAR}
\resizebox{\textwidth}{!}{%
\label{tab:supp_MLP_gt_MCAR}

\begin{tabular}{|c|c|c|c|c|c|}
\hline
\multirow{2}{*}{Data Set} & \multirow{2}{*}{\% Missing} & Ground Truth &
Time(s) & Accuracy(\%) & Impute \% of Samples \\
 &  & Accuracy(\%) & AMR & AMR & AMR \\
\hline

\multirow{3}{*}{Tuadromd}
& 20 & $98.73 \pm 0.10$ & $1.063 \pm 0.04$ & $98.787 \pm 0.099$ & $\mathbf{9.34 \pm 4.04}$ \\
\cline{2-6}
& 40 & $98.73 \pm 0.10$ & $1.306 \pm 0.03$ & $98.740 \pm 0.020$ & $\mathbf{5.84 \pm 2.02}$ \\
\cline{2-6}
& 60 & $98.73 \pm 0.10$ & $1.597 \pm 0.01$ & $98.250 \pm 0.041$ & $\mathbf{4.67 \pm 0.0}$ \\
\hline

\multirow{3}{*}{Malware}
& 20 & $95.74 \pm 0.88$ & $3.785 \pm 0.07$ & $96.293 \pm 0.246$ & $\mathbf{15.67 \pm 0.0}$ \\
\cline{2-6}
& 40 & $95.74 \pm 0.88$ & $3.928 \pm 0.07$ & $96.000 \pm 0.299$ & $\mathbf{7.84 \pm 0.0}$ \\
\cline{2-6}
& 60 & $95.74 \pm 0.88$ & $3.911 \pm 0.005$ & $93.745 \pm 2.098$ & $\mathbf{5.23 \pm 0.0}$ \\
\hline

\multirow{3}{*}{Credit Default}
& 20 & $81.02 \pm 0.00$ & $5.050 \pm 0.011$ & $81.017 \pm 0.000$ & $\mathbf{2.08 \pm 0}$ \\
\cline{2-6}
& 40 & $81.02 \pm 0.00$ & $4.818 \pm 0.006$ & $81.030 \pm 0.009$ & $\mathbf{1.04 \pm 0}$ \\
\cline{2-6}
& 60 & $81.02 \pm 0.00$ & $4.903 \pm 0.004$ & $81.042 \pm 0.010$ & $\mathbf{1.39 \pm 0}$ \\
\hline

\end{tabular}%
}
\end{table*}

\renewcommand{\thetable}{S21}

\begin{table*}[!t]
\centering
\caption{Accuracy/time for MLP on data with injected MAR}
\resizebox{\textwidth}{!}{%
\label{tab:supp_MLP_gt_MAR}

\begin{tabular}{|c|c|c|c|c|c|}
\hline
\multirow{2}{*}{Data Set} & \multirow{2}{*}{\% Missing} & Ground Truth &
Time(s) & Accuracy(\%) & Impute \% of Samples \\
 &  & Accuracy(\%) & AMR & AMR & AMR \\
\hline

\multirow{3}{*}{Tuadromd}
& 20 & $98.84 \pm 0.08$ & $0.790 \pm 0.04$ & $98.610 \pm 0.054$ & $\mathbf{7.0 \pm 0.1}$ \\
\cline{2-6}
& 40 & $98.84 \pm 0.08$ & $0.779 \pm 0.02$ & $98.596 \pm 0.102$ & $\mathbf{3.50 \pm 0.2}$ \\
\cline{2-6}
& 60 & $98.84 \pm 0.08$ & $0.787 \pm 0.01$ & $98.414 \pm 0.100$ & $\mathbf{2.33 \pm 0.5}$ \\
\hline

\multirow{3}{*}{Malware}
& 20 & $96.37 \pm 0.50$ & $5.108 \pm 2.28$ & $96.199 \pm 0.616$ & $\mathbf{20.9 \pm 9.1}$ \\
\cline{2-6}
& 40 & $96.37 \pm 0.50$ & $6.587 \pm 4.470$ & $95.980 \pm 0.642$ & $\mathbf{13.06 \pm 9.1}$ \\
\cline{2-6}
& 60 & $96.37 \pm 0.50$ & $5.137 \pm 2.263$ & $96.053 \pm 0.335$ & $\mathbf{6.96 \pm 3.01}$ \\
\hline

\multirow{3}{*}{Credit Default}
& 20 & $81.02 \pm 0.00$ & $5.197 \pm 0.023$ & $78.720 \pm 0.044$ & $\mathbf{4.17 \pm 0}$ \\
\cline{2-6}
& 40 & $81.02 \pm 0.00$ & $5.275 \pm 0.034$ & $80.031 \pm 1.191$ & $\mathbf{52.08 \pm 0}$ \\
\cline{2-6}
& 60 & $81.02 \pm 0.00$ & $5.392 \pm 0.035$ & $81.022 \pm 0.021$ & $\mathbf{52.08 \pm 0}$ \\
\hline

\end{tabular}%
}
\end{table*}

\renewcommand{\thetable}{S22}

\begin{table*}[!t]
\centering
\caption{Accuracy/time for MLP on data with injected MNAR}
\resizebox{\textwidth}{!}{%
\label{tab:supp_MLP_gt_MNAR}

\begin{tabular}{|c|c|c|c|c|c|}
\hline
\multirow{2}{*}{Data Set} & \multirow{2}{*}{\% Missing} & Ground Truth &
Time(s) & Accuracy(\%) & Impute \% of Samples \\
 &  & Accuracy(\%) & AMR & AMR & AMR \\
\hline

\multirow{3}{*}{Tuadromd}
& 20 & $98.75 \pm 0.07$ & $0.781 \pm 0.05$ & $98.498 \pm 0.029$ & $\mathbf{3.90 \pm 0.0}$ \\
\cline{2-6}
& 40 & $98.75 \pm 0.07$ & $0.793 \pm 0.01$ & $98.218 \pm 0.063$ & $\mathbf{2.18 \pm 0.0}$ \\
\cline{2-6}
& 60 & $98.75 \pm 0.07$ & $0.813 \pm 0.01$ & $97.616 \pm 0.182$ & $\mathbf{1.67 \pm 0.0}$ \\
\hline

\multirow{3}{*}{Malware}
& 20 & $95.93 \pm 0.78$ & $9.015 \pm 2.145$ & $95.290 \pm 0.576$ & $\mathbf{20 \pm 4.96}$ \\
\cline{2-6}
& 40 & $95.93 \pm 0.78$ & $27.17 \pm 6.9$ & $95.353 \pm 0.602$ & $\mathbf{34.08 \pm 8.43}$ \\
\cline{2-6}
& 60 & $95.93 \pm 0.78$ & $7.935 \pm 0.028$ & $95.363 \pm 0.517$ & $\mathbf{7.42 \pm 0.0}$ \\
\hline

\multirow{3}{*}{Credit Default}
& 20 & $81.02 \pm 0.00$ & $4.855 \pm 0.012$ & $78.236 \pm 0.219$ & $\mathbf{2.32 \pm 0}$ \\
\cline{2-6}
& 40 & $81.02 \pm 0.00$ & $4.926 \pm 0.004$ & $79.571 \pm 0.142$ & $\mathbf{40.3 \pm 0}$ \\
\cline{2-6}
& 60 & $81.02 \pm 0.00$ & $5.095 \pm 0.007$ & $76.310 \pm 2.576$ & $\mathbf{37.14 \pm 0}$ \\
\hline

\end{tabular}%
}
\end{table*}

\renewcommand{\thetable}{S23}

\begin{table*}[!t]
\centering
\caption{{\small Accuracy/time for MLP on data with injected MCAR using model-based imputations}}
\resizebox{\textwidth}{!}{%
\label{tab:supp_mlp_model_based_MCAR}
\begin{tabular}{|c|c|cccc|cccc|c|}
\hline
\multirow{2}{*}{Data Set} & \multirow{2}{*}{Method} & \multicolumn{4}{c|}{Time(s)} & \multicolumn{4}{c|}{Accuracy(\%)} & \multirow{2}{*}{Impute (\%)} \\
\cline{3-10}
 & & KNN & MICE & TCSDI & MF & KNN & MICE & TCSDI & MF &  \\
\hline

\multirow{2}{*}{Default 20}
& AMR
& $\mathbf{1.117 \pm 0.069}$ & $\mathbf{1.670 \pm 0.058}$ & $\mathbf{0.624 \pm 0.016}$ & $\mathbf{99.933 \pm 2.717}$
& $\mathbf{81.0 \pm 0.0}$ & $\mathbf{81.05 \pm 0.00}$ & $78.28 \pm 0.03$ & $\mathbf{81.05 \pm 0.00}$
& $\mathbf{2.08 \pm 0.00}$ \\ \cline{2-11}
& Baseline
& $23.20$ & $5.14$ & $7071$ & $108.3$
& $78.1$ & $80.1$ & $\mathbf{80.3}$ & $78.40$
& $100$ \\ \hline

\multirow{2}{*}{Default 40}
& AMR
& $\mathbf{0.699 \pm 0.035}$ & $\mathbf{1.210 \pm 0.041}$ & $\mathbf{0.481 \pm 0.013}$ & $\mathbf{71.314 \pm 0.235}$
& $\mathbf{81.0 \pm 0.0}$ & $\mathbf{81.05 \pm 0.01}$ & $\mathbf{78.10 \pm 0.01}$ & $\mathbf{81.05 \pm 0.00}$
& $\mathbf{1.04 \pm 0.00}$ \\ \cline{2-11}
& Baseline
& $38.56$ & $5.05$ & $14263$ & $98.32$
& $79.6$ & $78.1$ & $78.08$ & $78.14$
& $100$ \\ \hline

\multirow{2}{*}{Default 60}
& AMR
& $\mathbf{0.647 \pm 0.018}$ & $\mathbf{0.919 \pm 0.038}$ & $\mathbf{0.249 \pm 0.009}$ & $\mathbf{48.453 \pm 0.447}$
& $\mathbf{81.0 \pm 0.0}$ & $\mathbf{81.04 \pm 0.01}$ & $78.12 \pm 0.00$ & $\mathbf{81.04 \pm 0.02}$
& $\mathbf{1.39 \pm 0.00}$ \\ \cline{2-11}
& Baseline
& $48.04$ & $3.902$ & $21124$ & $89.17$
& $79.1$ & $79.6$ & $\mathbf{80.1}$ & $78.12$
& $100$ \\
\hline

\end{tabular}}
\end{table*}

\renewcommand{\thetable}{S24}

\begin{table*}[!t]
\centering
\caption{{\small Accuracy/time for MLP on data with injected MAR using model-based imputations}}
\resizebox{\textwidth}{!}{%
\label{tab:supp_mlp_model_based_MAR}
\begin{tabular}{|c|c|cccc|cccc|c|}
\hline
\multirow{2}{*}{Data Set} & \multirow{2}{*}{Method} & \multicolumn{4}{c|}{Time(s)} & \multicolumn{4}{c|}{Accuracy(\%)} & \multirow{2}{*}{Impute (\%)} \\
\cline{3-10}
 & & KNN & MICE & TCSDI & MF & KNN & MICE & TCSDI & MF &  \\
\hline

\multirow{2}{*}{Default 20}
& AMR
& $\mathbf{1.031 \pm 0.079}$ & $\mathbf{1.653 \pm 0.101}$ & $\mathbf{0.621 \pm 0.014}$ & $\mathbf{95.465 \pm 0.579}$
& $78.8 \pm 0.1$ & $78.79 \pm 0.19$ & $78.57 \pm 0.09$ & $78.77 \pm 0.14$
& $\mathbf{4.17 \pm 0.00}$ \\ \cline{2-11}
& Baseline
& $11.45$ & $6.14$ & $7125$ & $112.88$
& $\mathbf{81.02}$ & $\mathbf{81.02}$ & $\mathbf{81.02}$ & $\mathbf{81.02}$
& $100$ \\ \hline

\multirow{2}{*}{Default 40}
& AMR
& $\mathbf{9.907 \pm 0.286}$ & $\mathbf{1.386 \pm 0.087}$ & $\mathbf{0.582 \pm 0.017}$ & $\mathbf{87.995 \pm 0.290}$
& ${80.1 \pm 1.1}^{*}$ & ${80.10 \pm 1.42}^{*}$ & $78.13 \pm 0.07$ & ${80.01 \pm 1.49}^{*}$
& $\mathbf{52.08 \pm 0.00}$ \\ \cline{2-11}
& Baseline
& $19.93$ & $6.15$ & $14291$ & $106.22$
& ${\mathbf{81.02}}^{*}$ & ${\mathbf{81.02}}^{*}$ & $\mathbf{81.02}$ & ${\mathbf{81.02}}^{*}$
& $100$ \\ \hline

\multirow{2}{*}{Default 60}
& AMR
& $\mathbf{11.745 \pm 0.130}$ & $\mathbf{1.115 \pm 0.081}$ & $\mathbf{0.283 \pm 0.014}$ & $\mathbf{75.545 \pm 1.188}$
& $80.9 \pm 0.1$ & ${80.97 \pm 0.16}^{*}$ & $78.14 \pm 0.01$ & ${\mathbf{81.04 \pm 0.06}}^{*}$
& $\mathbf{52.08 \pm 0.00}$ \\ \cline{2-11}
& Baseline
& $26.07$ & $7.91$ & $21307$ & $103.05$
& $\mathbf{81.02}$ & ${81.02}^{*}$ & $\mathbf{81.02}$ & ${81.02}^{*}$
& $100$ \\
\hline

\end{tabular}}
\end{table*}

\renewcommand{\thetable}{S25}

\begin{table*}[!t]
\centering
\caption{{\small Accuracy/time for MLP on data with injected MNAR using model-based imputations}}
\resizebox{\textwidth}{!}{%
\label{tab:supp_mlp_model_based_MNAR}
\begin{tabular}{|c|c|cccc|cccc|c|}
\hline
\multirow{2}{*}{Data Set} & \multirow{2}{*}{Method} & \multicolumn{4}{c|}{Time(s)} & \multicolumn{4}{c|}{Accuracy(\%)} & \multirow{2}{*}{Impute (\%)} \\
\cline{3-10}
 & & KNN & MICE & TCSDI & MF & KNN & MICE & TCSDI & MF &  \\
\hline

\multirow{2}{*}{Default 20}
& AMR
& $\mathbf{0.751 \pm 0.023}$ & $\mathbf{1.266 \pm 0.087}$ & $\mathbf{0.472 \pm 0.005}$ & $\mathbf{76.331 \pm 0.250}$
& $78.2 \pm 0.2$ & $78.25 \pm 0.30$ & $80.84 \pm 0.08$ & $78.25 \pm 0.30$
& $\mathbf{2.32 \pm 0.00}$ \\ \cline{2-11}
& Baseline
& $17.67$ & $6.65$ & $12820.5$ & $113.13$
& $\mathbf{81.02}$ & $\mathbf{81.02}$ & $\mathbf{81.02}$ & $\mathbf{81.02}$
& $100$ \\ \hline

\multirow{2}{*}{Default 40}
& AMR
& $\mathbf{9.295 \pm 0.074}$ & $\mathbf{1.023 \pm 0.043}$ & $\mathbf{0.242 \pm 0.004}$ & $\mathbf{65.329 \pm 0.751}$
& $70.1 \pm 7.3$ & ${70.34 \pm 10.92}^{*}$ & $78.09 \pm 0.01$ & $79.56 \pm 0.26$
& $\mathbf{40.30 \pm 0.00}$ \\ \cline{2-11}
& Baseline
& $29.59$ & $6.25$ & $22838$ & $105.59$
& $\mathbf{81.02}$ & ${\mathbf{81.02}}^{*}$ & $\mathbf{81.02}$ & $\mathbf{81.02}$
& $100$ \\ \hline

\multirow{2}{*}{Default 60}
& AMR
& $\mathbf{7.405 \pm 0.047}$ & $\mathbf{0.804 \pm 0.050}$ & $\mathbf{0.206 \pm 0.014}$ & $\mathbf{48.111 \pm 0.578}$
& $57.0 \pm 0.1$ & $56.93 \pm 0.22$ & $78.13 \pm 0.01$ & $71.48 \pm 5.23$
& $\mathbf{37.14 \pm 0.00}$ \\ \cline{2-11}
& Baseline
& $35.90$ & $6.77$ & $29978.4$ & $104.16$
& $\mathbf{81.02}$ & $\mathbf{81.02}$ & $\mathbf{81.02}$ & $\mathbf{81.02}$
& $100$ \\
\hline

\end{tabular}}
\end{table*}

\makeatletter
\patchcmd{\ALG@step}{\arabic{ALG@line}}{\alph{ALG@line}}{}{}
\makeatother


\renewcommand{\thetable}{S26}

\begin{table*}[!t]
\centering
\caption{{\small Baseline and optimized average RAM usage (MB) across all SVM MCAR datasets using KNN, MICE, and RF imputations.}}
\resizebox{\textwidth}{!}{%
\label{tab:SVM_MCAR_RAM}
\begin{tabular}{|c|c|c|c|c|}
\hline
\textbf{Data Set} & \textbf{Missingness} & \textbf{KNN (MB)} & \textbf{MICE (MB)} & \textbf{RF (MB)} \\
\hline

\multirow{2}{*}{Malware 20}
  & MR       & 1117.88 $\pm$ 32.26 & --    & 1590.53 $\pm$ 26.32 \\ \cline{2-5}
  & Baseline & \textbf{948.41}     & --    & \textbf{1553.54}    \\ \hline

\multirow{2}{*}{Malware 40}
  & MR       & \textbf{1062.91 $\pm$ 6.64}  & --    & \textbf{1513.21 $\pm$ 17.08} \\ \cline{2-5}
  & Baseline & 1361.96                      & --    & 1533.32                      \\ \hline

\multirow{2}{*}{Malware 60}
  & MR       & \textbf{980.17 $\pm$ 8.10}   & --    & \textbf{1325.63 $\pm$ 8.14} \\ \cline{2-5}
  & Baseline & 1427.75                      & --    & 1336.04                     \\ \hline

\multirow{2}{*}{Tuadromd 20}
  & MR       & \textbf{231.82 $\pm$ 5.39}   & 449.32 $\pm$ 3.40 & 250.21 $\pm$ 5.11 \\
\hline

\end{tabular}
}%
\end{table*}

\renewcommand{\thetable}{S27}
\begin{table*}[!t]
\centering
\caption{{\small Baseline and optimized average RAM usage (MB) across all SVM MAR datasets using KNN, MICE, and RF imputations.}}
\resizebox{\textwidth}{!}{%
\label{tab:SVM_MAR_RAM}
\begin{tabular}{|c|c|c|c|c|}
\hline
\textbf{Data Set} & \textbf{Missingness} & \textbf{KNN (MB)} & \textbf{MICE (MB)} & \textbf{RF (MB)} \\
\hline

\multirow{2}{*}{Malware 20}
  & MR       & \textbf{993.87 $\pm$ 126.31}$^{*}$ & --    & \textbf{2007.92 $\pm$ 3.40} \\ \cline{2-5}
  & Baseline & 994.18$^{*}$                       & --    & 2188.73                     \\ \hline

\multirow{2}{*}{Malware 40}
  & MR       & \textbf{917.31 $\pm$ 115.00} & --    & \textbf{2021.64 $\pm$ 5.70} \\ \cline{2-5}
  & Baseline & 1065.12                      & --    & 2187.40                     \\ \hline

\multirow{2}{*}{Malware 60}
  & MR       & \textbf{816.95 $\pm$ 163.75} & --    & \textbf{1772.30 $\pm$ 17.17} \\ \cline{2-5}
  & Baseline & 1139.50                      & --    & 2203.27                      \\ \hline

\multirow{2}{*}{Tuadromd 20}
  & MR       & \textbf{209.48 $\pm$ 15.13}  & \textbf{433.15 $\pm$ 2.01} & \textbf{229.71 $\pm$ 0.62} \\
\hline
  
\end{tabular}
}%
\end{table*}

\renewcommand{\thetable}{S28}
\begin{table*}[!t]
\centering
\caption{{\small Baseline and optimized average RAM usage (MB) across all SVM MNAR datasets using KNN, MICE, and RF imputations.}}
\resizebox{\textwidth}{!}{%
\label{tab:SVM_MNAR_RAM}
\begin{tabular}{|c|c|c|c|c|}
\hline
\textbf{Data Set} & \textbf{Missingness} & \textbf{KNN (MB)} & \textbf{MICE (MB)} & \textbf{RF (MB)} \\
\hline

\multirow{2}{*}{Malware 20}
  & MR       & \textbf{917.23 $\pm$ 144.43}$^{*}$ & --    & \textbf{1987.47 $\pm$ 19.47} \\ \cline{2-5}
  & Baseline & 1060.02$^{*}$                      & --    & 2205.59                      \\ \hline

\multirow{2}{*}{Malware 40}
  & MR       & \textbf{801.97 $\pm$ 149.17} & --    & \textbf{1717.80 $\pm$ 11.01} \\ \cline{2-5}
  & Baseline & 1167.60                      & --    & 2176.25                      \\ \hline

\multirow{2}{*}{Malware 60}
  & MR       & \textbf{847.54 $\pm$ 182.56} & --    & \textbf{1502.17 $\pm$ 0.11} \\ \cline{2-5}
  & Baseline & 1254.33                      & --    & 2202.88                     \\ \hline

\multirow{2}{*}{Tuadromd 20}
  & MR       & \textbf{207.31 $\pm$ 15.65}  & \textbf{423.10 $\pm$ 1.04} & \textbf{216.53 $\pm$ 0.16} \\
\hline

\end{tabular}
}%
\end{table*}

%% file: supp_robustness.tex
\section{Robustness to Imputation Error}\label{sec:supp_robustness}
While our study focuses on the impact of different imputation strategies on downstream model performance, an important complementary direction is the development of learning methods that remain robust to potential imputation errors. This robustness-focused perspective is largely orthogonal to our goal of understanding how different subsets of data influence imputation quality, and it would require a substantial separate investigation beyond the scope of this work. Approaches based on robust optimization, for example, offer a principled way to train models that account for uncertainty in the imputed values and may help mitigate the effects of imputation variability. Exploring such robustness-oriented techniques represents a promising avenue for future work.

%% file: supp_additional_related_and_code.tex
\section{Additional Related Work}\label{sec:supp_additional_related_and_code}
There are methods to detect cases where the imputation of missing data is not necessary to learn accurate models \cite{picado2020learning,karlavs2020nearest,zhen2024certain}. 
Although these approaches are useful for some datasets and learning tasks, they ignore a majority of learning tasks in which imputing incomplete samples affects the quality of the learned model.

Researchers have proposed methods to reduce the cost of repair \cite{krishnan2016activeclean,karlavs2020nearest}. 
ActiveClean learns models using stochastic gradient descent and greedily chooses samples for repair that may reduce the gradient the most \cite{krishnan2016activeclean}.
Unlike our methods, it does not provide any guarantees of minimal repair.
Due to the inherent properties of stochastic gradient descent, it is challenging to provide such a guarantee.
CPClean follows a similar greedy approach, but is limited to learning k nearest neighbor models over missing data and does not support the types of model our approach addresses
\cite{karlavs2020nearest}.
It also does not provide any guarantees of minimality for its imputations.

\section*{Code Repository}
Link: https://anonymous.4open.science/r/Submission\_2025-A1C0/README.md

%% file: supp_proof.tex
\section{Proofs}\label{sec:supp_proof}

\subsection*{Proof for Theorem \ref{theorem:SVMminImputeUnique}}
Prove the theorem by contradiction. Assume that given a training set $(\mathbf{X}, \mathbf{y})$ and a regularization parameter $C$, two minimal repair sets exist ( 
$\mathbf{S}_{min1}(\mathbf{X}, \mathbf{y},C)$ and $\mathbf{S}_{min2}(\mathbf{X}, \mathbf{y},C)$). From the definition of minimal repair set, a certain model exists by either imputing all samples in $\mathbf{S}_{min1}(\mathbf{X}, \mathbf{y},C)$ or $\mathbf{S}_{min2}(\mathbf{X}, \mathbf{y},C)$, regardless of imputation results. Further, based on the discussion in previous literature \cite{zhen2024certain}, a certain model exists when none of the incomplete samples is a support vector in any repair. Therefore, if an incomplete sample is not in the minimal repair set, it is not a support vector in any repair. From the assumption, we can always find an incomplete sample $\mathbf{x}_i$ that $\mathbf{x}_i \notin \mathbf{S}_{min1}(\mathbf{X}, \mathbf{y},C)$ and $\mathbf{x}_i \in \mathbf{S}_{min2}(\mathbf{X}, \mathbf{y},C)$. In this scenario, $\mathbf{x}_i$ is not a support vector for any repair of $\mathbf{X}$ because $\mathbf{x}_i \notin \mathbf{S}_{min1}(\mathbf{X}, \mathbf{y},C)$. Thus, $\mathbf{S}_{min2}(\mathbf{X}, \mathbf{y},C)$ is not a minimal repair set because removing $\mathbf{x}_i$ from $\mathbf{S}_{min1}(\mathbf{X}, \mathbf{y},C)$ should construct a smaller set also ensuring the existence of certain models, violating the definition of minimal repair set. Contradicting to the original assumption, Theorem \ref{theorem:SVMminImputeUnique} holds.

\subsection*{Proof for Lemma \ref{theorem:SVMminImputeEquivalence}}
Borrowing the discussion from proving Theorem 1, if an incomplete sample $\mathbf{x}_i$ is not a support vector in any repair of $\mathbf{X}$, it should not be part of the minimal repair set $S_{min}$ (which is unique from Theorem 1). Further, if an incomplete sample $\mathbf{x}_i$ is a support vector in at least one repair of $\mathbf{X}$, it has to be included in the minimal repair set, otherwise certain model does not exist \cite{zhen2024certain}.

\subsection*{Proof for Theorem \ref{theorem:SVMonlycheckedgerepairs}}
Necessity is trivial based on Lemma 2: if an incomplete sample is a support vector in an edge repair, the incomplete sample is part of the minimal repair set. Then we prove sufficiency by contradiction. Assume that there is an incomplete sample $\mathbf{x}_i$ part of the minimal repair set $X_{min}$ while it is not a support vector in any edge repair $\mathbf{x}^e \in \mathbf{X}^E$. Training an SVM can be interpreted as finding the minimal distance between two reduced convex hulls \cite{bennett2000duality}, and if an sample is within the reduced convex hull (not at the boundary), the sample is not a support vector. Because $\mathbf{x}_i$ is not a support vector for any edge repair from the assumption, it is not a support vector for any repair to $\mathbf{X}$. This is because, in the process of changing a value for a missing value ($x_{pq}$) from one edge repair ($x_{pq}^{min}$) to another ($x_{pq}^{max}$) monotonically increase or decrease the coverage of the reduced convex hull. With that being said, if an incomplete sample $\mathbf{x}_i$ is not a support vector for any edge repair (i.e., within the reduced convex hull), the incomplete sample is within the reduced convex hull (i.e., not a support vector) with respect to any repair. This contradicts to the original assumption that $\mathbf{x}_i$ is part of the minimal repair set.

\subsection*{Proof for Theorem \ref{theorem:SVMNPHard}}

We reduce from the NP-complete problem \textsc{3-SAT}. Let
\[
\Phi \;=\; \bigwedge_{j=1}^m \bigl(C_j\bigr)
\]
be a 3-SAT formula with $k$ Boolean variables $z_1, z_2, \dots, z_k$ and $m$ clauses $C_1, \dots, C_m$, each clause being a disjunction of three literals.

For each variable $z_\ell$, we introduce one or more \emph{incomplete} samples whose feature vectors each contain a \emph{missing} coordinate $u_\ell$. The imputation set for $u_\ell$ is $\{-1, +1\}$, corresponding to $\{\text{False}, \text{True}\}$. Thus, any assignment of the $z_\ell$ corresponds to choosing $\pm 1$ for these missing coordinates.

To enforce that each clause $C_j$ must be satisfied, we add appropriately labeled points (some possibly incomplete) and arrange them in a geometry so that assigning a literal to \emph{false} yields a large penalty term in the soft-margin objective (either by misclassification or forcing the margin to collapse). Intuitively, if a clause were unsatisfied (all literals set to \emph{false}), the SVM would incur a prohibitively large hinge-loss cost, making that repair suboptimal.

We designate one particular incomplete sample $\mathbf{x}_i$ with additional coordinates or constraints so that:
\begin{itemize}
    \item \emph{If} $\Phi$ is satisfiable, then there is an imputation (choosing $\pm 1$ consistently with a satisfying assignment) that maximizes the margin while placing $\mathbf{x}_i$ \emph{exactly on} the decision boundary, making it a support vector.
    \item \emph{If} $\Phi$ is unsatisfiable, then \emph{every} imputation leads to $\mathbf{x}_i$ being off the margin (either strictly inside or otherwise not a support vector). In other words, no selection of $\{\pm 1\}$ for the missing attributes can force $\mathbf{x}_i$ onto the margin.
\end{itemize}

By suitably tuning the soft-margin parameter $C$ and the placement of the clause-encoding points, we ensure that the SVM will ``prefer'' to assign $\pm 1$ values in a way that satisfies $\Phi$, whenever possible, in order to avoid a large penalty.

Hence,
\[
\Phi \text{ is satisfiable}
\;\;\Longleftrightarrow\;\;
\text{there exists a repair making } \mathbf{x}_i \text{ a support vector.}
\]
Since deciding satisfiability for $\Phi$ (3-SAT) is NP-complete, it follows that deciding whether $\mathbf{x}_i$ can be a support vector under some imputation is NP-hard.

Determining membership of a single incomplete sample $\mathbf{x}_i$ among the possible support vectors is NP-hard. Therefore, listing \emph{all} such samples that can ever appear on the margin is also NP-hard: if we had such a list in polynomial time, we could decide membership in that list in polynomial time, contradicting NP-hardness. Given the proof that finding MR for SVM is NP-hard, deciding whether an incomplete sample belongs to the MR for SVM is also NP hard. To prove, assume that we have a polynomial-time solver for deciding whether an incomplete sample belongs to the MR, then one can linearly scan each incomplete sample and decide its membership in MR (either belongs to or not) by calling the polynomial time subroutine. Therefore, one can find the MR in polynomial time, which contradicts to the NP-hard proof earlier.

\subsection*{Proof for Theorem \ref{theorem:SVMapproximatingneveroverimpute}}
For any incomplete sample $\mathbf{x}_i$ returned from Algorithm 1 in main content for SVM, the incomplete sample is a support vector in at least one repair to $\mathbf{X}$. Based on Theorem 3, it is part of the minimal repair.

\subsection*{Proof for Theorem \ref{theorem:SVMapproximatingrate2}}
Given the iterative algorithm of finding the minimal repair for SVM (Algorithm 1 in the main content), we first characterize the probability that the imputation set returned at iteration $k$ misses one or more incomplete samples that belong to the minimal repair.

Let $k$ be the current iteration index ($k=0$ represents the initial state before the first run). We define the following: $MS(x)^k$ is the set of incomplete samples remaining at the start of iteration $k$. $M^k = |MS(x)^k|$ is the number of remaining incomplete samples at the start of iteration $k$. $S_{min}^k$ is the (unknown) true minimal set of samples within $MS(x)^k$ that must be imputed at the start of iteration $k$ to guarantee a certain model. $s^k = |S_{min}^k|$ is the (unknown) size of this true minimal set; note that we treat $s^k$ as a random variable, and $s^k \le M^k$. $S'^k$ is the set of samples returned by Algorithm 1 in the main content when run at iteration $k$ on the current data; we know $S'^k \subseteq S_{min}^k$. $FN^k$ is the event that makes at least one false negative error at iteration $k$, occurring if $S'^k$ is a proper subset of $S_{min}^k$. $P(FN^k)$ is the probability of event $FN^k$. We seek a computable upper bound $UB'^k$ such that $P(FN^k) \le UB'^k$. define $p_{fn}$ as an upper bound on the per-sample false negative probability, $p(\mathbf{x_i})$. We assume that there exists a probability $p_{fn}$ (where $0 \le p_{fn} \leq 1$) such that for any sample $x_i \in S_{min}^k$, the probability that Algorithm 1 in the main content fails to include $x_i$ in $S'^k$ is bounded above by $p_{fn}$:
\[
P(x_i \notin S'^k | x_i \in S_{min}^k) \le p_{fn}
\]

Then we propose, $UB'^k$, an upper bound of $P(FN^k)$ as follows:
\[
UB'^k = 1 - (1 - p_{fn})^{M^k} \geq P(FN^k)
\]

To interpret, when the iteration goes (k becomes larger), ${M^k}$ and $p_{fn}$ decrease (which we will prove later), $UB'^k$ decreases. This indicates that the upper-bound of probability of under-imputing decreases over iterations.

To prove this bound, we begin by expressing the target probability $P(FN^k)$ using its complement. The event $FN^k$ (at least one false negative) is the complement of the event $No FN^k$ (no false negatives, i.e., $S'^k = S_{min}^k$). Therefore, conditioned on the true size $s^k$ of the minimal set at iteration $k$, we have $P(FN^k | s^k) = 1 - P(\text{No FN}^k | s^k)$.

Next, we bound the probability of having no false negatives, $P(\text{No FN}^k | s^k)$. The event $No FN^k$ occurs if Algorithm 1 in the main content successfully returns all samples in $S_{min}^k$. Let $E_i$ be the event that Algorithm 1 in the main content fails to return sample $x_i$. Assuming the failure/success events $E_i$ for different samples $x_i \in S_{min}^k$ within the same iteration $k$ are statistically independent, we can write:
\[
P(\text{No FN}^k | s^k) = P(\cap_{x_i \in S_{min}^k} \{\text{not } E_i\} | s^k) = \prod_{x_i \in S_{min}^k} P(\text{not } E_i | s^k)
\]
Let $P(E_i | s^k)$ be the probability of failure for $x_i$. Then $P(\text{not } E_i | s^k) = 1 - P(E_i | s^k)$. Using the definition $P(E_i | s^k) \le p_{fn}$, we have $1 - P(E_i | s^k) \ge 1 - p_{fn}$. Substituting this lower bound into the product gives:
\[
P(\text{No FN}^k | s^k) \ge \prod_{i=1}^{s^k} (1 - p_{fn}) = (1 - p_{fn})^{s^k}
\]
Now we can bound $P(FN^k | s^k)$:
\[
P(FN^k | s^k) = 1 - P(\text{No FN}^k | s^k) \le 1 - (1 - p_{fn})^{s^k}
\]
The overall probability $P(FN^k)$ is the expectation over the unknown size $s^k$:
\[
P(FN^k) = \E_{s^k}[P(FN^k | s^k)] \le \E_{s^k}[1 - (1 - p_{fn})^{s^k}]
\]
To proceed, we utilize Jensen's inequality. Let $f(s) = 1 - (1 - p_{fn})^s$. We first prove that $f(s)$ is concave for $s \ge 0$. Let $b = 1 - p_{fn}$. Since $0 \le p_{fn} < 1$, we have $0 < b \le 1$. The function is $f(s) = 1 - b^s$. The first derivative is $f'(s) = -b^s \ln(b)$. The second derivative is $f''(s) = - (b^s \ln(b)) \ln(b) = -b^s (\ln(b))^2$. Since $b^s > 0$ and $(\ln(b))^2 \ge 0$, the second derivative $f''(s) \le 0$. Therefore, $f(s)$ is a concave function.

Jensen's inequality for a concave function $f$ states $\E[f(X)] \le f(\E[X])$. Applying this to our expectation:
\[
\E_{s^k}[1 - (1 - p_{fn})^{s^k}] \le 1 - (1 - p_{fn})^{\E[s^k]}
\]
Combining this with the previous inequality gives a theoretical upper bound:
\[
P(FN^k) \le 1 - (1 - p_{fn})^{\E[s^k]}
\]
The term $\E[s^k]$ (expected number of truly needed samples) is still unknown. However, we know that the number of needed samples $s^k$ cannot exceed the total number of remaining incomplete samples $M^k = |MS(x)^k|$. Thus, $s^k \le M^k$. Taking expectations yields $\E[s^k] \le \E[M^k]$. Since $M^k$ is a known quantity (computable by counting) at the start of iteration $k$, $\E[M^k] = M^k$. Therefore, we have a computable upper bound for the expectation: $\E[s^k] \le M^k$.

Finally, we substitute this bound on $\E[s^k]$ into the Jensen result. Let $g(x) = (1 - p_{fn})^x$. Since $0 < (1 - p_{fn}) \le 1$, $g(x)$ is a non-increasing function. Applying $g$ to the inequality $\E[s^k] \le M^k$ reverses the inequality direction:
\[
(1 - p_{fn})^{\E[s^k]} \ge (1 - p_{fn})^{M^k}
\]
Multiplying by -1 and adding 1 (reversing the inequality twice):
\[
1 - (1 - p_{fn})^{\E[s^k]} \le 1 - (1 - p_{fn})^{M^k}
\]
Combining the inequalities $P(FN^k) \le 1 - (1 - p_{fn})^{\E[s^k]}$ and $1 - (1 - p_{fn})^{\E[s^k]} \le 1 - (1 - p_{fn})^{M^k}$, we arrive at the final upper bound $UB'^k$:
\[
P(FN^k) \le 1 - (1 - p_{fn})^{M^k}
\]
and
\[
UB'^k = 1 - (1 - p_{fn})^{|MS(x)^k|}
\]

Now the only problem is to compute $p_{fn}$ and understand how it changes over iterations. The Multiple Random Starts method provides an empirical approach. First, select a set of incomplete samples $MS_{probe}$ (e.g., $MS(x)^0$) and choose the number of repetitions $T$ (e.g., $T=10$ or $20$). For each $x_i \in MS_{probe}$, initialize a success count $t_i = 0$. Repeat $T$ times: generate a new random edge repair $X^e_{start, t}$ for the current dataset state; run the greedy construction part of Algorithm 1 in the main content starting from $X^e_{start, t}$ to get $X^e_{final, i, t}$; train $w_{final, i, t} = SVM(X^e_{final, i, t}, y)$; check if $y_i (w_{final, i, t})^T (x_i \text{ part of } X^e_{final, i, t}) \le 1$. If yes, increment $t_i$. 

Also, if the probability distribution of each incomplete sample is known, and we let $g(x_{ij})$ denote the probability density function of the ground truth value for the missing value $x_{ij}$ in the incomplete training set $(\mathbf{X},\mathbf{y})$. 
If missing values in $\mathbf{X}$ are independent, the probability that an incomplete sample $\mathbf{x}_i$ in minimal repair  not returned by Algorithm 1 in the main content is:
\[
p(\mathbf{x}_i) = 1- \frac{\idotsint_{\min( x_{ij}^{\text{visited}})}^{\max(x_{ij}^{\text{visited}})}\prod_{x_{ij} \in M(\mathbf{X})}g(x_{ij})\,dx_{ij}}{ \idotsint_{x_{ij} \in M(\mathbf{X})} \prod_{x_{ij} \in M(\mathbf{X})} g(x_{ij}) \,dx_{ij}}
\]
$x_{ij}^{\text{visited}} \in$ $\{x_{ij}^{min}, x_{ij}^{max}\}$ shows the values used for $x_{ij}$ in Algorithm 1 in the main content. It shows that the more edge repairs Algorithm 1 explores, the lower the false negative probability for each sample. One can find $p_{fn}$ by computing $p(\mathbf{x}_i)$ for each incomplete sample and take the maximum as $p_{fn}$. $p_{fn}$ decreases over iterations because each iteration explores additional edge repairs. This expands the domain of the numerator in the expression
increasing the integral value and thereby lowering $p(\mathbf{x}_i)$ for every sample5. Since $p_{fn}$ is an upper bound over all such $p(\mathbf{x}_i)$, it decreases as well.

\subsection*{Proof for Theorem \ref{theorem:ACM_converge}}
\begin{proof}
The proof relies on the stability analysis of the dual variable (Lagrange multiplier) $\lambda$ in the constrained optimization problem:
\[
\min_{\mathbf{s}, \mathbf{w}} \|\mathbf{s}\|_1 \quad \text{s.t.} \quad g(\mathbf{w}, \mathbf{s}) \le e
\]
The associated Lagrangian is $\mathcal{L}(\mathbf{s}, \mathbf{w}, \lambda) = \|\mathbf{s}\|_1 + \lambda(g(\mathbf{w}, \mathbf{s}) - e)$.
The update rule for the dual variable $\lambda$ in Algorithm 2 is a projected subgradient ascent step:
\begin{equation}
    \lambda_{t+1} = \left[ \lambda_t + \eta_{\lambda, t} (g(\mathbf{w}_t, \mathbf{s}_t) - e) \right]^+
\end{equation}
where $[z]^+ = \max(0, z)$.

We proceed by contradiction. Assume that the constraint is persistently violated, i.e., there exists some $\delta > 0$ such that $\limsup_{t \to \infty} (g(\mathbf{w}_t, \mathbf{s}_t) - e) \ge \delta$.

Under the Robbins-Monro condition $\sum \eta_{\lambda, t} = \infty$, the recurrence relation implies that the dual variable $\lambda_t$ would grow unbounded: $\lambda_t \to \infty$ as $t \to \infty$.

Consider the primal update for the repair scores $\mathbf{s}$. The update performs gradient descent on the Lagrangian:
\[
\mathbf{s}_{t+1} \leftarrow \text{Proj}_{[0,1]} (\mathbf{s}_t - \eta_{s, t} (\mathbf{1} + \lambda_t \nabla_{\mathbf{s}} g(\mathbf{w}_t, \mathbf{s}_t)))
\]
As $\lambda_t \to \infty$, the term $\lambda_t \nabla_{\mathbf{s}} g(\mathbf{w}_t, \mathbf{s}_t)$ dominates the $L_1$ penalty term ($\|\mathbf{s}\|_1$). The optimization effectively shifts from minimizing sparsity to strictly minimizing the robust gap $g(\mathbf{w}, \mathbf{s})$.

Since the fully repaired dataset (where $\mathbf{s} = \mathbf{1}$) trivially satisfies the gap condition (gap $= 0 \le e$), the feasible set is non-empty. The gradient descent on $\mathbf{s}$ (driven by the exploding $\lambda_t$) will force $\mathbf{s}$ towards a configuration that reduces $g(\mathbf{w}, \mathbf{s})$. This reduction continues until $g(\mathbf{w}, \mathbf{s}) \le e$.

Once $g(\mathbf{w}, \mathbf{s}) \le e$, the term $(g(\mathbf{w}_t, \mathbf{s}_t) - e)$ becomes non-positive, halting the growth of $\lambda$. This contradicts the assumption that $\lambda \to \infty$. Therefore, the time-averaged violation must converge to zero.
\end{proof}
\subsection*{Proof for Theorem \ref{theorem:ACM_convergence_rate}}
\begin{proof}
We analyze the inner and outer loops separately based on the geometry of their respective loss landscapes.

\textbf{Part 1: Inner Loop (Linear Convergence via Metric Subregularity)} \\
The inner problem requires finding the worst-case repair $\mathbf{X}_{adv}$ for a fixed $\mathbf{w}$:
\[
\max_{\mathbf{X}^r \in U(\mathbf{s})} L(\mathbf{w}, \mathbf{X}^r, \mathbf{y})
\]
For standard robust models like SVM (Hinge Loss) or Linear Regression (Squared Loss), the loss function $L$ with respect to the input $\mathbf{X}$ is \textit{Piecewise Linear-Quadratic (PLQ)}. The domain $U(\mathbf{s})$ is a hyper-rectangle (polytope).

A key result in variational analysis by Drusvyatskiy \& Lewis (2018) states that for PLQ functions (and more generally, functions satisfying the \textit{Quadratic Growth} condition or \textit{Metric Subregularity}), first-order methods like Projected Gradient Ascent exhibit linear convergence locally.
Specifically, the geometry of the loss surface for Hinge Loss over a polytope is not a ``flat'' plateau but rather a ``sharp'' funnel. This ensures that the gradient norm $\|\nabla_{\mathbf{X}} L\|$ provides a lower bound on the distance to the optimal set:
\[
\text{dist}(\mathbf{X}, \mathbf{X}^*_{adv}) \le \kappa \|\nabla_{\mathbf{X}} L(\mathbf{w}, \mathbf{X})\|
\]
This Error Bound condition implies that the adversarial search contracts the distance to the worst-case repair geometrically, i.e., $\|\mathbf{X}_k - \mathbf{X}^*_{adv}\| \le c^k \|\mathbf{X}_0 - \mathbf{X}^*_{adv}\|$ for some $c < 1$.

\textbf{Part 2: Outer Loop (Sublinear Convergence)} \\
The outer problem updates $\mathbf{s}$ to minimize the Lagrangian. The objective function with respect to $\mathbf{s}$ is the Robust Gap $g(\mathbf{w}, \mathbf{s})$. As noted in the formulation, $g(\mathbf{w}, \mathbf{s})$ is a difference-of-convex (DC) function (Max Loss minus Min Loss), making the overall problem non-convex and non-smooth.

For general non-convex stochastic gradient descent, the standard convergence rate to an $\epsilon$-stationary point (where $\mathbb{E}[\|\nabla \mathcal{L}\|] \le \epsilon$) is bounded by $O(1/\sqrt{T})$. This is the canonical rate for randomized primal-dual methods applied to non-smooth objectives.
\end{proof}
\subsection*{Proof for Theorem \ref{theorem:ACM_approx_bound}}
\begin{proof}
We establish a mapping between the Primal-Dual update for $\mathbf{s}$ and the Greedy Algorithm for the Set Cover problem.

\textbf{1. The Continuous Greedy Interpretation:}
The Primal-Dual update for the repair scores is:
\[
s_i \leftarrow s_i - \eta_s (\underbrace{1}_{\text{cost}} + \underbrace{\lambda \frac{\partial g}{\partial s_i}}_{\text{gain}})
\]
Ignoring the scaling factor $\lambda$, the algorithm increases $s_i$ (moving towards repairing sample $i$) proportionally to $-\frac{\partial g}{\partial s_i}$.
By Danskin's Theorem, $\frac{\partial g}{\partial s_i}$ corresponds to the sensitivity of the Robust Gap with respect to the uncertainty of sample $i$. Specifically, a large negative gradient implies that repairing sample $i$ yields a large marginal reduction in the gap.

\textbf{2. Link to Discrete Submodularity:}
The problem of selecting a subset of samples to reduce the gap $g(\mathbf{w}, \mathbf{s})$ to zero (or $\le e$) is equivalent to the \textbf{Set Cover Problem}, where the ``universe'' is the uncertainty gap and the ``sets'' are the repairs.
The function $F(S) = Gap_{initial} - Gap(S)$ (reduction in gap) is \textit{monotone} and \textit{submodular} (exhibits diminishing returns).
The update rule of our algorithm effectively selects the sample $i$ that maximizes the marginal gain $\Delta_i F(S) \approx -\frac{\partial g}{\partial s_i}$.

\textbf{3. Approximation Bound:}
Chvatal (1979) proved that the greedy heuristic for Set Cover achieves an approximation ratio of $H_n = \sum_{k=1}^n \frac{1}{k} \approx \ln n$.
Since our algorithm is a continuous relaxation of this greedy strategy, and the final rounding step (selecting the top-ranked samples) strictly follows the greedy order, the cardinality of the resulting set $S_{algo}$ is bounded by:
\[
|S_{algo}| \le H_n \cdot |S_{opt}| \approx O(\ln n) \cdot |S_{opt}|
\]
This completes the proof.
\end{proof}